\documentclass[10pt,journal,compsoc]{IEEEtran}

\usepackage{booktabs}
\usepackage{lipsum}

\usepackage{amsmath}
\usepackage{bm,cuted}
\usepackage[ruled,vlined,linesnumbered]{algorithm2e}
\usepackage{graphicx}
\usepackage{tabularx} 
\usepackage{amssymb}
\usepackage[safe]{tipa}
\usepackage{amsthm}
\usepackage{multirow}
\usepackage{subfigure}
\usepackage{xcolor}
\usepackage{dsfont}
\usepackage{upgreek}
\usepackage{url}
\newcommand{\etal}{\textit{et al.}}

\usepackage{pifont}
\newcommand{\cmark}{\ding{51}}%
\newcommand{\xmark}{\ding{55}}%
\makeatletter
\newcommand\HUGE{\@setfontsize\Huge{20}{30}}
\makeatother

\def\eqref#1{equation~\ref{#1}}

\def\1{\bm{1}}

\DeclareMathAlphabet{\mathsfit}{\encodingdefault}{\sfdefault}{m}{sl}
\SetMathAlphabet{\mathsfit}{bold}{\encodingdefault}{\sfdefault}{bx}{n}

\ifCLASSOPTIONcompsoc
  \usepackage[nocompress]{cite}
\else
  \usepackage{cite}
\fi

\ifCLASSINFOpdf
 
\else

\fi

\title{Style Transfer for 2D Talking Head Animation}

\author{Binh X. Nguyen$^{1}$, Tuong Do$^{1}$, Hien Nguyen$^{1}$, Vuong Pham$^{1}$, Toan Tran$^{3}$,\\Erman Tjiputra$^{1}$, Quang Tran$^{1}$, Anh Nguyen$^{2}$\\
{$^{1}$AIOZ, Singapore}\\
{$^{2}$University of Liverpool, UK}\\
{$^{3}$VinAI Research}\\
{\tt\small \{binh.xuan.nguyen,tuong.khanh-long.do,hien.nguyen,vuong.pham,erman.tjiputra,quang.tran\}}\\
{\tt\small @aioz.io}\\
{\tt\small anh.nguyen@liverpool.ac.uk}\\
{\tt\small v.toantm3@vinai.io}}

\author{Trong-Thang Pham, Nhat Le, Tuong Do, Hung Nguyen, Erman Tjiputra, Quang D. Tran, Anh Nguyen
\thanks{Trong-Thang Pham, Nhat Le, Tuong Do, Hung Nguyen, Erman Tjiputra, Quang D. Tran are with AIOZ, Singapore {\{thang.trong.pham, nhat.minh.le, tuong.khanh-long.do, hung.nguyen.manh, erman.tjiputra, quang.tran\}@aioz.io}.}
\thanks{Anh Nguyen is with Department of Computer Science, University of Liverpool, UK (email: anh.nguyen@liverpool.ac.uk).}
\thanks{Two first authors contribute equally.}
\thanks{Corresponding author: Anh Nguyen.}
}

\begin{document}

\IEEEtitleabstractindextext{

\begin{abstract}

Audio-driven talking head animation is a challenging research topic with many real-world applications. Recent works have focused on creating photo-realistic 2D animation, while learning different talking or singing styles remains an open problem. In this paper, we present a new method to generate talking head animation with learnable style references. Given a set of style reference frames, our framework can reconstruct 2D talking head animation based on a single input image and an audio stream. Our method first produces facial landmarks motion from the audio stream and constructs the intermediate style patterns from the style reference images. We then feed both outputs into a style-aware image generator to generate the photo-realistic and fidelity 2D animation. In practice, our framework can extract the style information of a specific character and transfer it to any new static image for talking head animation. The intensive experimental results show that our method achieves better results than recent state-of-the-art approaches qualitatively and quantitatively. Our source code can be found at: \url{https://github.com/aioz-ai/AudioDrivenStyleTransfer}.

\end{abstract}

}

\maketitle
\IEEEdisplaynontitleabstractindextext

\IEEEpeerreviewmaketitle

\section{Introduction}

Talking head animation is an active research topic in both academia and industry. This task has a wide range of real-world interactive applications such as digital avatars~\cite{cudeiro_capture_2019}, speech tutoring~\cite{dey2010talking}, video conferencing~\cite{wang_one-shot_2021}, virtual reality~\cite{morishima1998real,badin1998towards,latif2021talking}, computer games~\cite{xie2015expressive,guo2021adneft}, and digital animations~\cite{aiozGdance}. Given an arbitrary input audio and a 2D image (or a set of 2D images) of a character, the goal of talking head animation is to generate photo-realistic frames. 
The output can be the 2D~\cite{zhou2020makelttalkMIT,guo2021adneft,lu2021liveLSP} or 3D talking head~\cite{cudeiro_capture_2019,zhou_visemenet_2018,taylor_deep_2017}. 
With recent advances in deep learning, especially generative adversarial networks~\cite{goodfellow_generative_2014}, several works have addressed  different aspects of the talking head animation task such as head pose control~\cite{zhou2021pose,zhang20213d}, facial expression~\cite{le2022global,le2023uncertainty}, emotion generation~\cite{livingstone2018ryersonRAVDESS,eskimez2021speech}, and photo-realistic synthesis~\cite{zhou2020makelttalkMIT,vougioukas2019end,chen_photo-realistic_2019}.

While there has been considerable advancement in the generation of talking head animation, achieving photo-realistic and fidelity animation is not a trivial task. It is even more challenging to render natural motion of the head with different styles~\cite{cudeiro_capture_2019}. In practice, several aspects contribute to this challenge. First, generating a photo-realistic talking head using only a single image and audio as inputs requires multi-modal synchronization and mapping between the audio stream and facial information~\cite{edwards2016jali}. In many circumstances, this process may result in fuzzy backgrounds, ambiguous fidelity, or abnormal face attributes~\cite{zhou2020makelttalkMIT}. 
Second, various talking and singing styles can express diverse personalities~\cite{walker1997improvising}. Therefore, the animation methods should be able to adapt and generalize well to different styles~\cite{walker1997improvising}. Finally, controlling the head motion and connecting it with the full-body animation remains an open problem~\cite{jiang2016real}.
 
\begin{figure}[ht]
\centering
\includegraphics[width=0.5\textwidth]{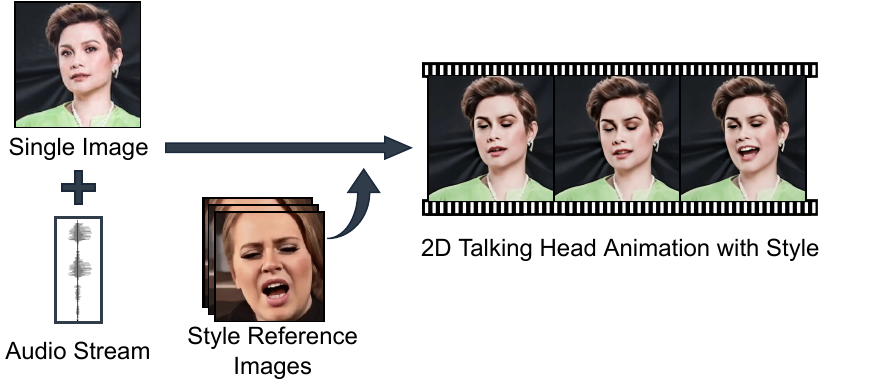}
\label{fig:introvis}
\caption{Given an audio stream, a single image, and a set of style reference frames, our method generates realistic 2D talking head animation. Please also see our supplementary video. Video \textit{Adele \copyright YouTube} (public domain). Image \textit{Lea \copyright Twitter} (public domain).}\vspace{3ex}
\end{figure}

Recently, several methods have been proposed to generate photo-realistic talking heads~\cite{guo2021adneft,lu2021liveLSP,suwajanakorn_synthesizing_2017,zhou2020makelttalkMIT} or to match the pose from a source video~\cite{zhou2021pose} while little work has focused on learning the personalized character style~\cite{lu2021liveLSP}. In practice, apart from personalized talking style, we have different singing styles such as \texttt{ballad} and \texttt{rap}. These styles pose a more challenging problem for talking head animation as they have the unique eye, head, mouth, and torso motion. The facial movements of singing styles are also more varied and dynamic than the talking style. Therefore, learning and bringing these styles into 2D talking heads is more challenging. Currently, most of the style-aware talking head animation methods do not fully disentangle the audio style information and the visual information, which causes ambiguity during the transferring process~\cite{lu2021liveLSP}.

In this work, we present a new deep learning framework called Style Transfer for 2D talking head animation. Our framework provides an effective way to transfer talking or singing styles from the style reference to animate single 2D portrait of a character given an arbitrary input audio stream. We first generate photo-realistic 2D animation with natural expression and motion. We then propose a new method to transfer the personalized style of a character into any talking head with a simple style-aware transfer process. 
Figure~\ref{fig:introvis} shows an overview of our approach.

In summary, our contributions are as follows:
\begin{itemize}
    \item We propose a new framework for generating photo-realistic 2D talking head animations from the audio stream as input. 
    \item We present a style-aware transfer technique, which enables us to learn and apply any new style to the animated head. Our generated 2D animation is photo-realistic and high fidelity with natural motions.
    
    \item We conduct intensive analysis to show that our proposed method outperforms recent approaches qualitatively and quantitatively. Our source code and trained models will be released for reproducibility.
\end{itemize}

\section{Literature Review}
\label{Sec:Literature}

\textbf{2D Talking Head Animation.} 
Creating talking head animation from an input image and audio has been widely studied in the past few years. 
One of the earliest works~\cite{bregler_video_1997} considered this as a sorting task that reorders images from footage video using the phoneme sequence. 
Based on~\cite{bregler_video_1997},~\cite{garrido_vdub_2015} proposed to capture 3D model from dubber and actor in order to synthesize photo-realistic face.~\cite{ezzat_trainable_2004} introduced a learning approach to create a trainable system that could synthesize a mouth shape from an unseen utterance. 
Later works focused on audio-driven to generate realistic mouth shapes~\cite{suwajanakorn_synthesizing_2017,taylor_deep_2017} or realistic faces~\cite{zhou2019talking,songtalking,chen2019hierarchical,mittal2020animating, grassal2022neural}. 
The authors in~\cite{eskimez_generating_2018} focused on generating full facial landmarks using the input audio. ~\cite{wiles_x2face_2018} moved into a different direction by focusing on generating talking face that include pose and expression of another face video.
Instead of creating talking face,~\cite{greenwood_joint_2018} designed a model that produces head motion from the joint latent space using BiLSTM.~\cite{zakharov2019few,zakharov2020fast,chen2020talking,kumar2020robust,liang2022expressive} pave the way for creating realistic head avatars.~\cite{vougioukas2019end, vougioukas_realistic_2020,wangaudio2head,ren2021pirenderer,zhang2021flow,hong2022depth,wang2022one} used only a single image and audio to develop an end-to-end generation network. 
\cite{chen2019hierarchical} focused on handling noise and different facial shapes and angles. 
\cite{ginosar_learning_2019} proposed a model that can learn conversational gestures. 
\cite{yin2022styleheat,yao2022dfa,liu2022semantic,hong2022headnerf} focused on generating fidelity talking head which natural head pose and photo-realistic motions. Recently, \cite{lu2021liveLSP,zhou2022dialoguenerf} proposed to generate photo-realistic talking head with personalized information encoded.

\begin{figure*}[ht]
    \centering
    \includegraphics[width=\textwidth, keepaspectratio=true]{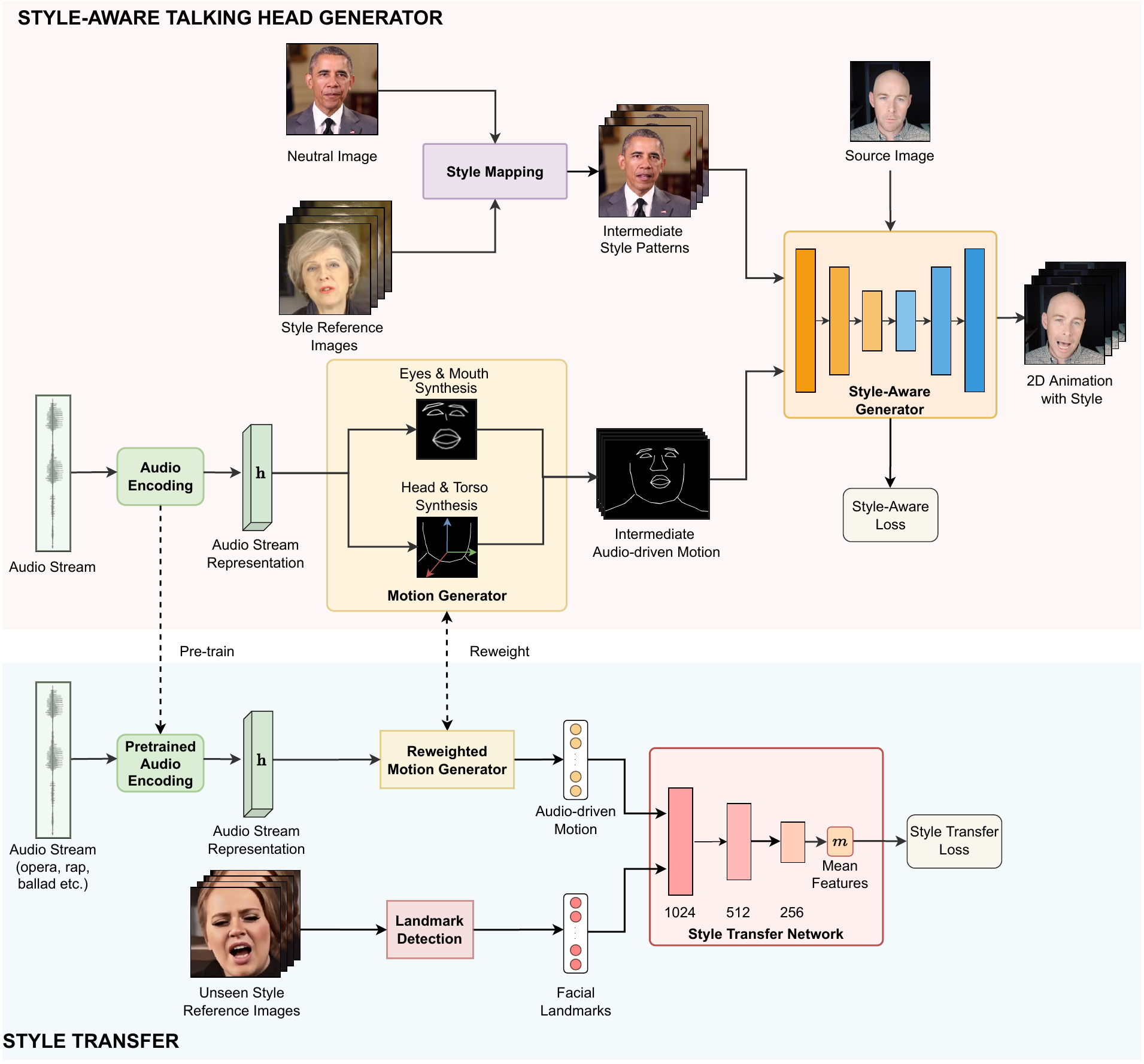}
    \vspace{-3ex}
    \caption{A detailed illustration of our method.}
    \vspace{-1ex}
    \label{fig:overview}
\end{figure*}

\textbf{Speaker Style Estimation.} There are many kinds of speaker styles such as generic, personal, controlled pose, or special expression. Generic style could be learned by training on multiple videos \cite{zhou2020makelttalkMIT,thies_neural_2020,richard_meshtalk_2021}, while personalized style could be decided by training on one avatar particularly~\cite{suwajanakorn_synthesizing_2017, lu2021liveLSP, lahiri_lipsync3d_2021}. In~\cite{zhou2021pose}, the authors introduced a method that generates controllable poses with an input video.~\cite{wiles_x2face_2018} transferred poses and expressions from another video input.~\cite{garrido_vdub_2015} mapped the style from dubber to actor.~\cite{siarohin2019first, romero2021smile, wang_one-shot_2021} captured motions from the driven video and transferred them into input image during the generation process.~\cite{tsao2009ensemble} tried to ensemble speaker and speaking environment to characterize the speaker variability in the environment.~\cite{liu2015video} leveraged a pre-captured database of 3D mouth shapes and associated speech audio from one speaker to refine the mouth shape of a new actor.
Likewise, many works did not restrict to a specific style but could be adapted to general and controllable results~\cite{zhou_visemenet_2018,cudeiro_capture_2019}.

\textbf{Speech Representation for Face Animation.} Some prior works used hand-crafted models to match phoneme and mouth shape in each millisecond audio signal as speech representation~\cite{bregler_video_1997, zhou_visemenet_2018}. DeepSpeech~\cite{hannun_deep_2014} paved the way for learning a speech recognition system using an end-to-end deep network. Following that, ~\cite{greenwood_joint_2018} trained Deep Bi-Directional LSTMs to learn a language-long-term structure that model the relationship between speech and the complex activity of faces.
~\cite{suwajanakorn_synthesizing_2017} used Mel-frequency spectral coefficients to synthesize high-quality mouth texture of a character, and then combined it with a 3D pose matching method to synchronize the lip motion with the audio in the target animation.
In our work, similar to~\cite{lu2021liveLSP}, we use manifold learning to generalize the style information from speech representation.

\section{Preliminaries}
\subsection{Audio Encoding}
The input audio provides critical input information to our system.
Following~\cite{lu2021liveLSP}, we use auto-regressive predictive coding (APC)~\cite{chung_generative_2020} to extract structural audio stream representations. Given historical data, the APC model uses the $\log$ Mel spectrograms feature as input and predicts future surface properties. The model is a $3$-layer GRU~\cite{2014ChoGRU}.

\begin{equation}
    \label{eq:speed_represent}
    \mathbf{h}_u = \rm{GRU}^{(u)}(\mathbf{h}_{u - 1}), \forall u \in [1,3] 
\end{equation}
where $\mathbf{h}_u \in \mathbb{R}^{512}$ represents the posterior probability of every layer in GRUs. Our targeted audio stream representations are latent states in the final GRU unit. During the training process, we add a linear layer that maps the output to predict the next log Mel spectrogram. 
The representation of the final GRU unit $\mathbf{h}_3$ is projected to a manifold as in~\cite{lu2021liveLSP} to extract the audio stream feature $\mathbf{h} \in \mathbb{R}^{512}$. In practice, this projection step can improve the generalization of the audio stream extractor~\cite{lu2021liveLSP}.

\subsection{Motion Generator}
\label{subsubsec:motionReconstruction}
Given the extracted audio features, this step generates audio-driven motions in our framework. In practice, the character's style is mainly defined by the mouth, eye, head, and torso movement. Therefore, we consider the motion around these regions of the face in our work.

\textbf{Mouth and Eye Synthesis.} There has been a lot of work towards predicting mouth movements from audio. Our goal is to learn a mapping from acoustic information to the intermediate representation using a deep network.
While most other methods apply parameters of a parametric model~\cite{chen_photo-realistic_2019,taylor_deep_2017}, 3D vertices~\cite{cudeiro_capture_2019}, or facial blend shapes~\cite{thies_neural_2020}, we employ 3D displacements (i.e., the geometry of mouth and eyes surfaces)~\cite{zhou2020makelttalkMIT,lu2021liveLSP} in object coordinates relative to the target character's locations as the intermediate representation.

In practice, we use three-stacked LSTM layers~\cite{1997LSTM} with $256$ units, accompanied by three multilayer perceptrons (MLP) as the learning network. Each MLP layer has $256$, $512$, and $75$ neurons, subsequently. Note that, following~\cite{lu2021liveLSP}, we add $18$ frames delay to make the LSTM model robust to a short future. The three-stacked LSTM network is trained to predict the  3D displacements sequence $\left(\Delta\hat{\mathbf{v}}_{1}, \Delta\hat{\mathbf{v}}_{2}, \ldots, \Delta\hat{\mathbf{v}}_{T} \right)$, with $\Delta\hat{\mathbf{v}}_{t} \in \mathbb{R}^{41 \times 3}$, given the ground truth sequence $\left(\Delta\mathbf{v}_{1},\Delta\mathbf{v}_{2}, \ldots, \Delta\mathbf{v}_{T} \right)$. The loss $\mathcal{L}_{\rm me}$ for the mouth and eye synthesis is the Euclidean distance between the ground truth sequence and the predicted sequence:
\begin{equation} \mathcal{L}_{\rm  me} = 
    \sum_{t=1}^{T}\left\|\Delta \mathbf{v}_{t}-\Delta \hat{\mathbf{v}}_{t}\right\|_{\rm f}^{2},
\end{equation}
where $T=240$ as in~\cite{lu2021liveLSP} and $\rm f$ is the Frobenius norm.

\textbf{Head and Torso Motion.}
We train a conditional probabilistic generative network~\cite{oord2016conditional} to learn the head pose distribution. This network generates new head pose $\mathbf{x}_t \in \mathbb{R}^{6} $ at timestamp $t$. The first three elements of $\mathbf{x}_t$ are the rotation vector
, and the last three elements are the translation vector. Then, given the head pose sequence
($\mathbf{x}_{t-255}$, $\ldots$,  $\mathbf{x}_{t-1}$) and the audio stream feature $\mathbf{h}_{t}$, the loss $\mathcal{L}_{\rm ht}$ for the head and torso motion is defined as follow:
\begin{equation}
    \mathcal{L}_{\rm ht} = -\ln \left(\mathcal{N}\left(\mathbf{x}_{t}, \mathbf{h}_{t} \mid \mu_{\rm x}, \varepsilon_{\rm x}\right)\right)
    \label{eq_loss_gaussian}
\end{equation}
where $\mathbf{x}_{t}, \mathbf{h}_{t}$ is the input head pose and the audio feature at time $t$. Using Equation~\ref{eq_loss_gaussian}, the model predicts the mean values $\mu_{\rm x}$ and standard deviations $\varepsilon_{\rm x}$ of the Gaussian distribution of the input. 

The Motion Generator loss $\mathcal{L}_{\rm mg}$ is a sum of $\mathcal{L}_{\rm me}$ for mouth/eyes synthesis and $\mathcal{L}_{\rm ht}$ for the head/torso motion:

\begin{equation}
\mathcal{L}_{\rm mg}= \mathcal{L}_{\rm me} + \mathcal{L}_{\rm ht}
\end{equation}

\section{Style-Aware Talking Head Generator}
\label{sec:Methodology}
Our goal is to generate high fidelity 2D talking head animation while allowing personalized style transfer. To achieve this, we first disentangle the style information encoded in Style Reference Images. The style is then synthesized with a neural network to generate Intermediate Style Pattern. Along with the Intermediate Audio-driven Motion produced by the Motion Generator, the Intermediate Style Pattern and the source image are passed into the Style-Aware Generator to generate the 2D talking head with style. Figure~\ref{fig:overview} shows the details of our style-aware 2D talking head generation method.

\subsection{Style Reference Images} 
\label{subsec:styleref}
To learn the character's styles more effectively, we define the Style Reference Images as a set of images retrieved from a video of a specific character by using the key motion templates. 
Inspired by~\cite{lu2021liveLSP},~\cite{zhou2020makelttalkMIT}, and music theory about rhythm~\cite{arvaniti2009rhythm}, we use four key motion templates that contain popular motion range and behavior. 
Each behavior is then plotted as a reference style pattern, which is used to retrieve the ones that are most similar in each video in the dataset. 
To retrieve similar patterns, we apply similarity search~\cite{chen2021augnet} for each image in the video of the character. The result image set is called the Style Reference Images and is used to provide character's styles information in our framework.

\subsection{Style Mapping}
\label{subsec:Reenactment}
The Style Mapping is designed to disentangle the style in the reference images and then map the extracted style to the neutral image.
Then, the input of this module is a pair of two images: a neutral image $\textbf{\textit{I}}_{s}$, and a style reference image $\textbf{\textit{I}}_{r}$. The output is an Intermediate Style Pattern (ISP - an image) which has the identity that comes from $\textbf{\textit{I}}_{s}$ and the style represented in $\textbf{\textit{I}}_{r}$. ISP has the visual information of the neutral image but the style is from the style reference image. In practice, we first disentangle the style information encoded in the pose and expression of both the neutral and reference image, then map the style from the reference image into the neutral image to generate the output ISP image $\textbf{\textit{I}}_{o}$.

\textbf{Disentangling Neutral Image.}
Since the head pose, expression, and keypoints from the neutral image contain the style information of a specific character, they need to be disentangled to learn the style information. In this step, given an input image $\textbf{\textit{I}}_{s}$, a set of $\rm{k}$ number of keypoints $\rm{c}_k$ is disentangled first to store the geometry signature by a Keypoint Extractor network. Then, we extract the pose, parameterized by a translation vector $\tau \in \mathbb{R}^3$ and a rotation matrix $\rm{R} \in \mathbb{R}^{3\times 3}$, and expression information $\varepsilon_k$ from the image by a Pose Expression network. 
After the disentangling process, we can reconstruct the image keypoints $\rm{C}_k$ using Equation~\ref{eq:source_kpts_gen}.
The extracted keypoints maintain the geometry signature and style information of the head in the neutral image. 
\begin{equation}
    \label{eq:source_kpts_gen}
    \rm{C}_k = \rm{c}_k \times \rm{R} +\tau + \varepsilon_k
\end{equation}

\textbf{Disentangling Style Reference Image.}
Similar to the neutral image, we use two deep networks to disentangle and extract the head pose and keypoints from the style reference image. However, instead of extracting new keypoints from the reference images, we reuse the extracted ones $\rm{c}_k$ from the neutral image, which contains the identity-specific geometry signature of the neutral image. The final keypoints $\bar{\rm{C}}_k$ of the style reference image are computed in Equation~\ref{eq:drive_kpts_gen}:

\begin{equation}
    \label{eq:drive_kpts_gen}
    \bar{\rm{C}}_k = \rm{c}_k \times \bar{\rm{R}} +\bar{\tau} + \bar{\varepsilon}_k
\end{equation} 
where $\bar{\tau} \in \mathbb{R}^3$, $\bar{\rm{R}} \in \mathbb{R}^{3\times 3}$ and $\bar{\varepsilon}$ are translation vector, rotation matrix, and expression information extracted from the style reference image, respectively.

\textbf{Style Mapping.} To construct the Intermediate Style Pattern $\textbf{\textit{I}}_{o}$, 
we first extract two keypoints sets $\rm{C_k}$ and $\bar{\rm{C}}_k$ from the neutral image and the style reference image. We then estimate the warping function based on the two keypoints sets to warp the encoded features of the source (neutral image) to the target so that it can represent the style of the reference image. 
Then, we feed the warped version of the source encoded features and the extracted style information into an Intermediate Generator to obtain the ISP image. In practice, we choose the neutral image as a general image in Obama Weekly Address dataset~\cite{suwajanakorn_synthesizing_2017}, while the style reference image is one of the four images in the Style Reference Images set. By applying the style mapping process for all four images in the Style Reference Images, we obtain a set of four ISP images. This set (the Intermediate Style Pattern - ISP) is used as the input for the Style-Aware Generator in Section~\ref{subsec:style-aware-translation}.

We note that the Style Mapping is necessary to obtain the ISP because we want the model pays attention to the pose/expression of a reference style, not the identity-specific visual information of a character. In practice, without loss of generality, we choose Obama's facial images as the canonical representation to map the style information.
This aspect is important for the Style Transfer process in Section~\ref{sec:onTheFlyFineTuning} as we want the model to effectively learn and synthesize facial motion from the input audio without depending on any specific character.

\subsection{Style-Aware Generator}
\label{subsec:style-aware-translation}

This module generates a 2D talking head from a source image, the generated intermediate motion, and the style information represented in the Intermediate Style Pattern. In this module, the facial map plays an essential role in explicitly identifying groups of facial keypoints, which makes the style-aware learning process easier to converge. We note that the ISP is not the facial map but images of identity-neutral representation obtained from the Style Mapping. During the training, the Style-Aware Generator has not been re-weighted by any specific styles through the Style Transfer process (Section~\ref{sec:onTheFlyFineTuning}), hence the generated talking head has neutral style.

\begin{figure}[ht]
    \centering
    \includegraphics[width=0.3\textwidth, keepaspectratio=true]{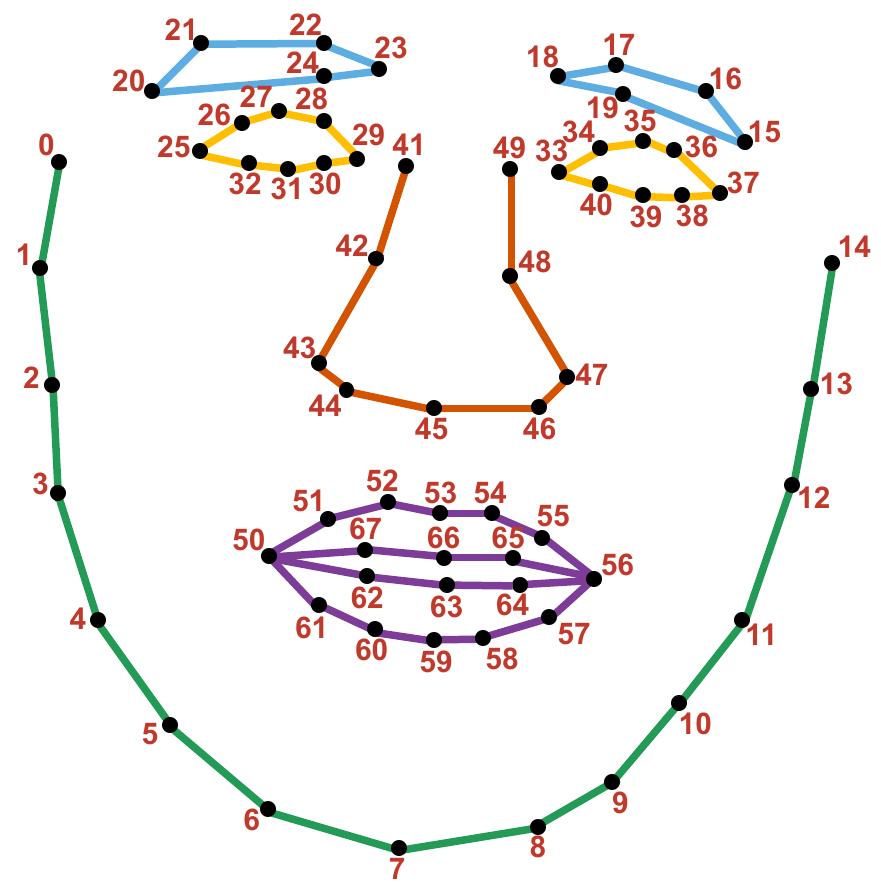}
    \vspace{0.2ex}
    \caption{Facial map with keypoints are semantically placed in groups.}
    \label{fig:partMap}
\end{figure}

\textbf{Facial Map.} The concept of the facial map is to limit the learning space within groups of keypoints between motions generated by Motion Generator and motions represented by ISP. By constructing a facial map with motion keypoints, we can mark and plot keypoints of different parts of motions that need to be focused on during the Style Transfer phase. In our experiment, the facial map has the size of $512 \times 512$ and can be obtained by connecting consecutive keypoints in a preset semantic sequence and projecting it onto the 2D image plane using a pre-computed camera matrix. Our pre-defined facial map is shown in Figure~\ref{fig:partMap}.

\textbf{Network Architecture.}
In Style-Aware Generator, we design our training process as an adversarial scheme~\cite{goodfellow_generative_2014}. Our network consists of a generator $\mathbf{G}$ that aims at generating images to fool the discriminator $\mathbf{D}$. For the discriminator $\mathbf{D}$'s backbone, we use PatchGAN~\cite{isola_image--image_pix2pix2017, wang_high-resolution_2018}. For the generator $\mathbf{G}$, we use encoder-decoder architecture with skip connections~\cite{ronneberger2015uUnet,lu2021liveLSP}. In particular, the generator $\mathbf{G}$ consists of 8 convolutional layers. The output of each layer contains ($256^2$, $128^2$, $64^2$, $32^2$, $16^2$, $8^2$, $4^2$, $2^2$) pixels, and the corresponding number of channels is ($64$, $128$, $256$, $512$, $512$, $512$, $512$, $512$). Each layer has a stride of $2$ and a residual block, except for the first layer. The corresponding symmetric decoder layer is similar to the encoder layer.

\textbf{Adversarial Loss.}
We optimize the discriminator $\mathbf{D}$ by using LSGAN loss \cite{mao_least_2017}:
\begin{equation}
    \mathcal{L}_{\mathbf{D}}=(y'_\mathbf{D}-1)^{2}+y_\mathbf{D}^{2}
\end{equation}
where $y_\mathbf{D}, y'_\mathbf{D}$ is the output of the discriminator $\mathbf{D}$ when we use the ground truth image $\textbf{\textit{I}}$ and the generated image $\textbf{\textit{I}}'$ as the input.  

The generator loss $\mathcal{L}_{\mathbf{G}}$ is the combination of the following losses:
\begin{itemize}
    \item An adversarial loss $\mathcal{L}_{\rm A}=(y_\mathbf{D}-1)^{2}$ introduced by~\cite{mao_least_2017} to encourage the realism of the generated images.
    \item A $L_{1}$ pixel wise loss $\mathcal{L}_{\rm{pw}}=\|\textbf{\textit{I}}-\textbf{\textit{I}}'\|_{1}$ to minimize differences at pixel-wise level.
    \item  A perceptual loss $\mathcal{L}_{\rm P}$~\cite{johnson_perceptual_2016} to minimize high-level differences, i.e., content and style discrepancies.
    \item  A feature matching loss $\mathcal{L}_{\rm F}$ introduced in~\cite{lu2021liveLSP} to minimize differences at the feature level.
\end{itemize}

We then compute the generator loss  $\mathcal{L}_{\mathbf{G}}$ as:
\begin{equation}
    \mathcal{L}_{\mathbf{G}}=\mathcal{L}_{\rm A}+\lambda_{\rm{pw}} \mathcal{L}_{\rm{pw}}+\lambda_{\rm P} \mathcal{L}_{\rm P}+\lambda_{\rm F} \mathcal{L}_{\rm F},
\label{eq:feat_match_loss}
\end{equation}
where $\lambda_{\rm pw}, \lambda_{\rm P}, \lambda_{\rm F}$  are hyper-parameters to control the contribution of each loss term.

\textbf{Style-Aware Loss.} 
The Intermediate Style Patterns in Section~\ref{subsec:Reenactment} are expected to comprehensively carry the key characteristics of one person. However, the Style-Aware Generator may not be aware of these style patterns and fail to generate desired results. 
In practice, we also find that using only the loss $\mathcal{L}_{\mathbf{G}}$ may fail in some cases with special styles such as rapping or opera singing.
To address this problem, we further introduce the style-aware photometric loss $\mathcal{L}_{\rm{sp}}$. This loss is combined with the generator loss $\mathcal{L}_{\mathbf{G}}$ to improve the generation quality and penalize the generated output that has a high deviation from the reference style patterns. The style-aware photometric loss is formulated as the pixel-wise error between the generated image $\textbf{\textit{I}}'$ and the matched style pattern image $\textbf{\textit{I}}_{\rm {m}}$:

\begin{equation}
    \mathcal{L}_{\rm{sp}}=\Vert \mathbf{W} \odot (\textbf{\textit{I}}' - \textbf{\textit{I}}_{\rm {m}}) \Vert_{1}
    \label{eq_style_aware_loss}
\end{equation}
where $\mathbf{W}$ is the weighting mask which has values depending on different face regions; $\odot$ denotes the Hadamard product; the matched style pattern image $\textbf{\textit{I}}_{\rm {m}}$ is obtained by using~\cite{chen2021augnet} to retrieve the best-matched image corresponding to one of the style reference images.
To acquire $\mathbf{W}$, we first use an off-the-shelf face parsing method to generate the segmentation mask of the face~\cite{lin2021faceparse}. To achieve high fidelity image generation, we want the network to focus more on each facial region. Specifically, the corresponding weight of $\mathbf{W}$ according to mouth, eyes, and skin regions are set to  $5.0, 3.0, 1.0$, respectively. Note that weights for other regions in the weighting mask $\mathbf{W}$, e.g. background, are set to $0$.

\section{Style Transfer}
\label{sec:onTheFlyFineTuning}

The style transfer phase focuses on transferring the styles to a new character by re-weighting the Motion Generator given the input audio. In our transferring phase, we assume that the talking or singing styles are encoded in both the audio stream and reference images.
Therefore, this style information is learnable and can be transferred from one to another character. As in~\cite{ahuja2020style}, we mainly rely on the pre-trained models from the training phase to perform the style transfer. Since Style-Aware Generator can cover the visual information generated from different styles, our goal in this phase is to make sure the style encoded in the Intermediate Audio-driven Motions can be adjusted to different styles rather than just the neutral one (i.e., the styles in the training data). We capture both the audio stream and reference images as the input in this stage. See Figure~\ref{fig:overview} for the details of our style transfer process. 

Given the reference images and an audio stream (e.g., \texttt{opera}, \texttt{rap}, etc.), we first use the pre-trained audio encoding to extract the audio feature and apply the Motion Generator to reconstruct the audio-driven motion $\bm{\phi}_{\rm mg}$. The reference images are fed through a pre-trained landmark detector to extract theirs corresponding facial landmarks $\bm{\phi}_{\rm s}$. The generated motions and facial landmarks are vectorized into $(68 \times 3)$-dimensional vectors. Both $\bm{\phi}_{\rm mg}$ and $\bm{\phi}_{\rm s}$ are then passed through a style transfer network to extract the mean features. A style transfer loss $\mathcal{L}_{\rm transfer}$ is then optimized through back-propagation. The mean features are the latent encoded vector containing both information from the audio-driven landmarks and the facial landmarks.
 
\subsection{Style Transfer Network} The style transfer network $f(\cdot)$ aims to learn the differences between motions of the input reference images and audio-driven motions extracted from the Motion Generator. Thanks to the style transfer loss $\mathcal{L}_{\rm transfer}$, the network is optimized to lower the gap of both mentioned motions, and then re-weight the parameters of Motion Generator to generate output motions that is similar to the target style. 
After re-weighting, the Motion Generator can produce style-aware audio-driven motions which are then passed into Style-Aware Generator to generate 2D animation with style. The style transfer network has three multilayer perceptrons (MLP), each MLP layer has $1024$, $512$, and $256$ neurons, subsequently. The final layer produces the mean features used in the style transfer loss.

\subsection{Style Transfer Loss} The style transfer loss is proposed to assure the generated motions take into account the target style. This loss is in-cooperated with the Motion Generator loss $\mathcal{L}_{\rm mg}$ for fine-tuning the Motion Generator module during transferring process. 
The style transfer loss $\mathcal{L}_{\rm transfer}$ is contributed by the constraint loss $\mathcal{L}_{\rm sc}$ and the regularization loss $\mathcal{L}_{\rm r}$. The constraint loss is introduced to learn the style from the source motion and then transfer it into the generated one through the style transfer network.  
\begin{equation} 
   \label{eq:styleconst_loss} 
   \mathcal{L}_{\rm sc} = \left\lVert f(\bm{\phi}_{\rm{mg}}) - f(\bm{\phi}_{\rm s}) \right\lVert_2^{2} 
\end{equation} 
where $f(\cdot)$ is the style transfer network. 

The regularization loss $\mathcal{L}_{\rm r}$ aims to increase the generalization of the style transfer process. Besides, it can deal with extreme cases of the generated motions that may break the manifold of valid styles and negatively affect the generated images. This loss is computed as:
\begin{equation}
    \label{eq:stylereg_loss}
    \mathcal{L}_{\rm r} = \bigg(\left\lVert \nabla_{\bm{\hat\phi}_{\rm{mg}}}f(\bm{\hat\phi}_{\rm{mg}}) \right\lVert_2 - 1\bigg)^{2}
\end{equation}
where $\bm{\hat\phi}_{\rm{mg}}$ is the joint representation that controls the contribution of source motion $\bm{\phi}_{\rm s}$ during the style learning process.  $\bm{\hat\phi}_{\rm{mg}}$ is computed from $\bm{\phi}_{\rm s}$ and $\bm{\phi}_{\rm{mg}}$ as follows:
\begin{equation}
    \label{eq:style_const}
    \bm{\hat\phi}_{\rm {mg}} = \gamma \bm{\phi}_{\rm s} + (1 - \gamma) \bm{\phi}_{\rm {mg}}
\end{equation}
where $\gamma$ controls the amount of leveraged style information.

The final transferring loss $\mathcal{L}_{\rm {transfer}}$ is computed as:

\begin{equation}
\mathcal{L}_{\rm {transfer}} = \mathcal{L}_{\rm mg} + \mathcal{L}_{\rm sc} + \mathcal{L}_{\rm r}    
\end{equation}

So as to control the style, both reference images and the audio stream are required during the transferring process.

\section{Implementation} 
\label{sec:ImplementNDatasetAcq} 
  
\subsection{Data Processing} 
\label{subsec:dataProcessing} 
\textbf{Style Reference Images.} 
To learn the styles from different speakers, we require an audio-visual dataset with a broad selection of speakers to learn the speaker-aware dynamics variations of head motion and facial expressions. We identified that the VoxCeleb2~\cite{chung2018voxceleb2} dataset is ideal for our needs because it comprises video snippets from a wide range of speakers.
Since our purpose is to capture speaker dynamics for talking head synthesis, we picked a subset of $67$ speakers from VoxCeleb2 with a total of $1,232$ video clips.  
We have about $5-10$ minutes of footage for each speaker. For each video, we perform image retrieval to find key motion frames and use them as the Style Reference Images. During retrieval, we use each motion template from 4 pre-defined key motion templates (mentioned in Section~\ref{subsec:styleref}) to retrieve from a series of motion maps in the reference clip. Then, we collect indexes and obtain corresponding images and audio signals, which are expected to have the style information.

\textbf{Data Processing.} 
All videos from the VoxCeleb2~\cite{chung2018voxceleb2} dataset are extracted at $60$ FPS. 
We first trim the video to retain the face in the center, then resize it to $512\times 512$. Our internal face tracker is leveraged to obtain $68$ key points on the face. Face segmentation~\cite{lin2021faceparse} is used to obtain the skin mask.  
Following~\cite{lu2021liveLSP}, the head and torso motion is manually identified for the first frame of each series and tracked for the remaining frames using optical flow.  
  
For illustration purposes, we use the following images, videos, and audios in our experiments: \textit{May \copyright UK Government} (open government license); \textit{Mac \copyright Genius} (public domain); \textit{Andrea \copyright Houston Symphony} (public domain); \textit{Adele \copyright YouTube} (public domain); \textit{Obama \copyright Barack Obama Foundation} (public domain); \textit{Nadella \copyright IEEE Computer Society} (public domain); \textit{McStay \copyright Darren McStay} (CC BY); \textit{Trump \copyright White House} (open government license); \textit{Lea \copyright Twitter} (public domain); \textit{Emma \copyright L'avenir} (CC BY); \textit{Natalie \copyright Facebook} (public domain); \textit{Scarlett \copyright YouTube} (public domain); 
\textit{Mona Lisa \copyright Twitter} (public domain). \textit{Easy on me}~\cite{EasyOnMe}, \textit{One minutes rap}~\cite{Rap}, \textit{Opera – The Ultimate Collection}~\cite{Opera}.
  
\subsection{Training} 
\label{subsec:DataPreparation} 
\textbf{Audio Encoding.} Following~\cite{lu2021liveLSP, zhou2020makelttalkMIT}, we use the Common Voice dataset~\cite{ardila2020common} to train the Audio Encoder. 
There are around $26$ hours of unlabeled statements throughout all samples.
Note that $80$-dimensional log Mel spectrograms are employed as surface representation and are computed with $\frac{1}{120}$(s) frame-shift, $\frac{1}{60}$(s) frame length, and $512$-point STFT~\cite{griffin1984signal} representation.

\textbf{2D Head Generator.} 
For generating 2D talking heads, there are three modules that need to be trained, including Style Mapping, Motion Generator, and Style-Aware Generator. 
To train the Style Mapping module, we use the input image from the Obama Weekly Address dataset~\cite{suwajanakorn_synthesizing_2017} and the style of different characters from Style Reference Images. The Style Mapping will extract the Intermediate Style Pattern sets of different videos in VoxCeleb2~\cite{chung2018voxceleb2} which are further used as the input of the Style-Aware Generator. 
Both the Motion Generator and Style-Aware Generator are jointly trained using the VoxCeleb2~\cite{chung2018voxceleb2} dataset accompanied by corresponding Intermediate Style Pattern for each selected video. 
We also use a best-estimated affine transformation~\cite{segal2009generalized} to register the facial landmarks obtained by the Data Processing step in Section~\ref{subsec:dataProcessing} to a front-facing standard facial template. As a result, the speaker-dependent head pose is factored out.

\subsection{Style Transfer}
During the style transfer process, we first utilize the style transfer network pre-trained on VoxCeleb2~\cite{chung2018voxceleb2}. Then, we fine-tune the style transfer network with the learning rate of $10^{-3}$ and freeze other modules in the first epoch. After that, we unfreeze all modules and fine-tune the whole network for $5$ epochs with a learning rate of $10^{-7}$. Finally, the Cosine Annealing~\cite{loshchilov2016sgdr} is used as a learning rate scheduler during the style transfer process.

\subsection{Implementation Details}

We implement our framework using PyTorch. We train the network on the NVIDIA Titan V GPU with Adam optimizer~\cite{kingma2014adam}. The learning rate is set to $10^{-4}$, $10^{-4}$, $10^{-5}$, $10^{-4}$ to train the Audio Encoding, the Motion Generator, the Style-Aware Generator, and the Style Mapping, respectively. The batch size is set to $8$ for the Style-Aware Generator and $64$ for other modules.
The hyper-parameters $\lambda_{\rm pw}, \lambda_{\rm P}, \lambda_{\rm F}$ in Equation~\ref{eq:feat_match_loss} are set to $(100,10,1)$ based on validation results.
\section{Results}
\subsection{Qualitative Evaluation}

\subsubsection{Capacity Analysis}
We first qualitatively compare the capacity of our proposed method with recent approaches using four different criteria: 2D Photo-realistic, One-shot Synthesis, Style Learning, and Style Transfer. Here is the description of each criterion:

\begin{itemize}
    \item 2D Photo-realistic:  Indicate the visual fidelity and detailness of the generated output in each method.
    \item One-shot Synthesis: Indicate if the method inputs a single 2D image to generate the talking head or not.
    \item Style Learning: Indicate the method's ability to learn the person-specific style of a particular character. 
    \item Style Transfer: Indicate the method's ability to transfer a particular character style to a new target. 
\end{itemize}

\begin{table}[!ht]
\centering
\caption{Method capacity comparison between our method and recent approaches.}
\resizebox{\linewidth}{!}{
\setlength{\tabcolsep}{0.15 em} 
{\renewcommand{\arraystretch}{1.2}
\begin{tabular}{c|c|c|c|c}
\hline
\multirow{3}{*}{\textbf{Methods}} & \multicolumn{4}{c}{\textbf{Criteria}}  \\ \cline{2-5} 
                                   & \multirow{2}{*}{\textbf{\begin{tabular}[c]{@{}c@{}}2D Photo-\\ Realistic\end{tabular}}} & \multirow{2}{*}{\textbf{\begin{tabular}[c]{@{}c@{}}One-shot\\ Synthesis\end{tabular}}} & \multirow{2}{*}{\textbf{\begin{tabular}[c]{@{}c@{}}Style\\ Learning\end{tabular}}} & \multirow{2}{*}{\textbf{\begin{tabular}[c]{@{}c@{}}Style\\ Transfer\end{tabular}}}\\
                             & & & &\\\hline
MIT~\cite{zhou2020makelttalkMIT}     & \xmark                             & \cmark                                    & \xmark                              & \xmark                                \\ \hline
AudioDVP~\cite{wen2020photorealistic}     & \cmark                             & \cmark                                    & \xmark                              & \xmark                                \\ \hline
PCT~\cite{zhou2021pose} & \xmark                              & \cmark                                    & \cmark                              & \xmark                                 \\ \hline
AD-NERF~\cite{guo2021adneft} & \cmark                              & \xmark                                    & \xmark                              & \xmark                                \\ \hline
FACIAL~\cite{zhang2021facial}     & \cmark                              & \xmark           & \xmark                          & \xmark                                                             \\ \hline
LSP~\cite{lu2021liveLSP}     & \cmark                              & \xmark           & \cmark                          & \xmark                                                             \\ \hline
Ours     & \cmark                           & \cmark                                    & \cmark                              & \cmark                                \\ \hline
\end{tabular}
}}
\vspace{2ex}

\vspace{-2ex}
\label{tab:ad_method_compare}
\end{table}

\begin{figure*}[ht] 
   \centering
  \huge
\resizebox{\linewidth}{!}{
\setlength{\tabcolsep}{2pt}
\begin{tabular}{cccccccccc}

\rotatebox[origin=l]{90}{\hspace{-0.3cm} \textbf{[Vougioukas et. al.]}} &
\shortstack{\includegraphics[width=0.33\linewidth]{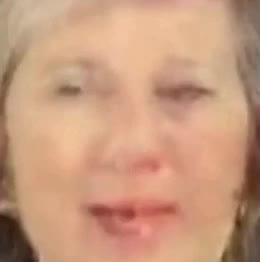}}&
\shortstack{\includegraphics[width=0.33\linewidth]{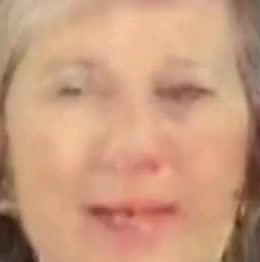}}&
\shortstack{\includegraphics[width=0.33\linewidth]{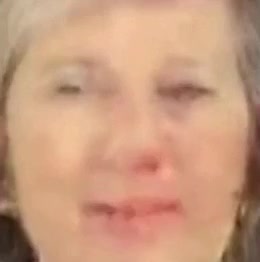}}&
\shortstack{\includegraphics[width=0.33\linewidth]{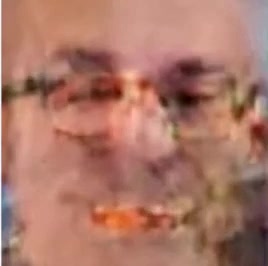}}&
\shortstack{\includegraphics[width=0.33\linewidth]{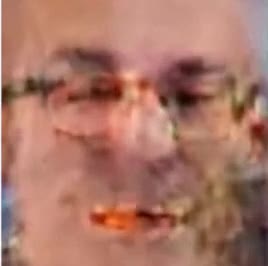}}&
\shortstack{\includegraphics[width=0.33\linewidth]{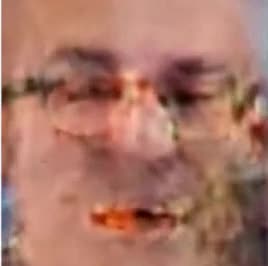}}&
\shortstack{\includegraphics[width=0.33\linewidth]{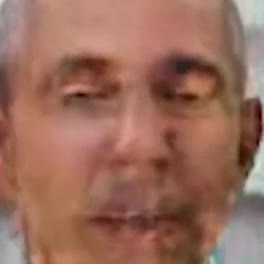}}&
\shortstack{\includegraphics[width=0.33\linewidth]{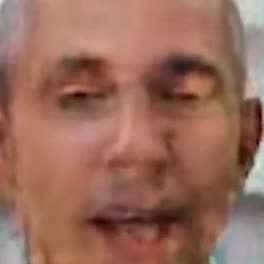}}&
\shortstack{\includegraphics[width=0.33\linewidth]{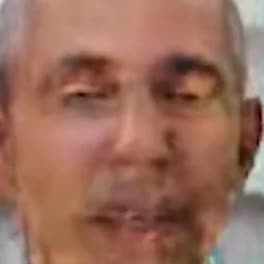}}\\[1pt]
\rotatebox[origin=l]{90}{\hspace{0.4cm}  \textbf{[Chen et. al.]}} &
\shortstack{\includegraphics[width=0.33\linewidth]{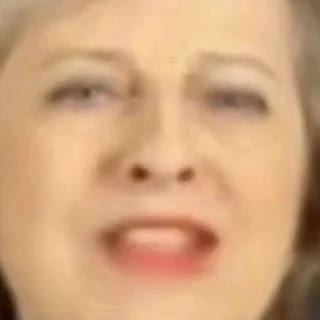}}&
\shortstack{\includegraphics[width=0.33\linewidth]{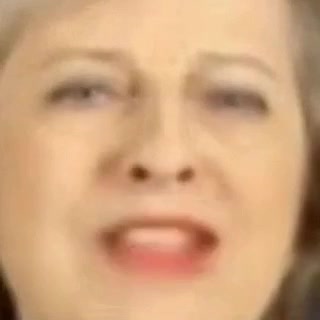}}&
\shortstack{\includegraphics[width=0.33\linewidth]{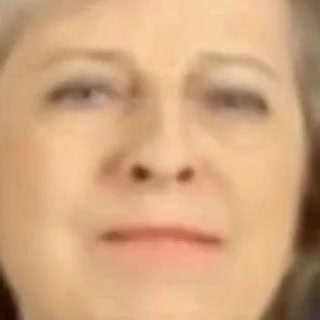}}&
\shortstack{\includegraphics[width=0.33\linewidth]{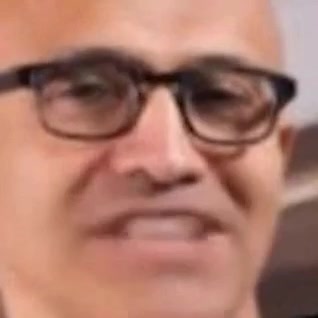}}&
\shortstack{\includegraphics[width=0.33\linewidth]{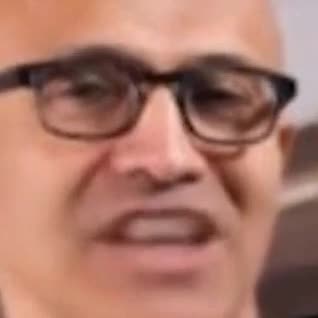}}&
\shortstack{\includegraphics[width=0.33\linewidth]{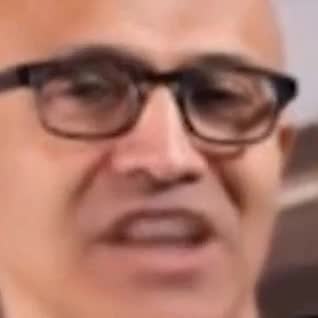}}&
\shortstack{\includegraphics[width=0.33\linewidth]{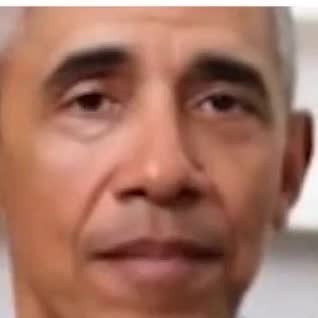}}&
\shortstack{\includegraphics[width=0.33\linewidth]{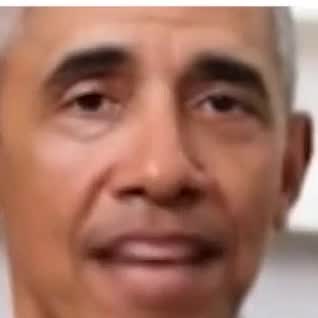}}&
\shortstack{\includegraphics[width=0.33\linewidth]{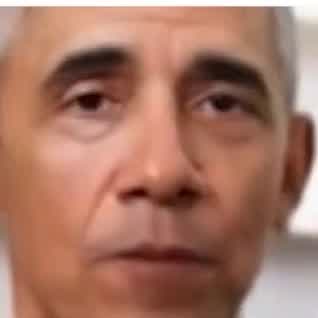}}\\[1pt]
\rotatebox[origin=l]{90}{\hspace{0.5cm}  \textbf{[Zhou et. al.]}} &
\shortstack{\includegraphics[width=0.33\linewidth]{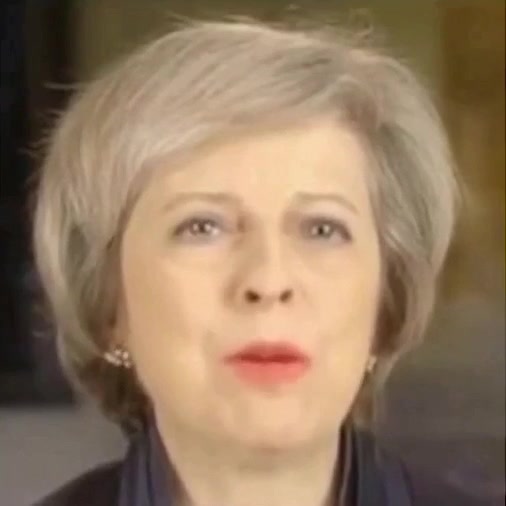}}&
\shortstack{\includegraphics[width=0.33\linewidth]{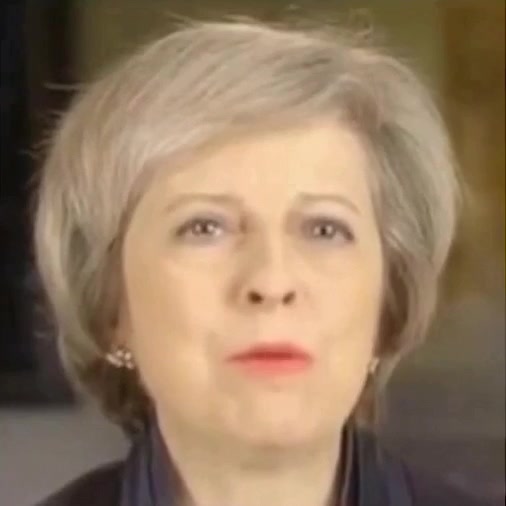}}&
\shortstack{\includegraphics[width=0.33\linewidth]{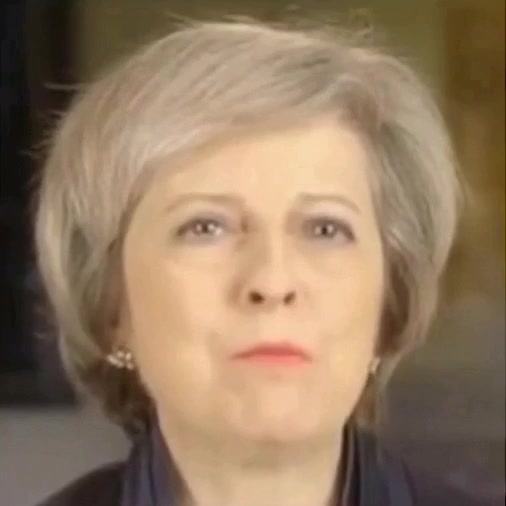}}&
\shortstack{\includegraphics[width=0.33\linewidth]{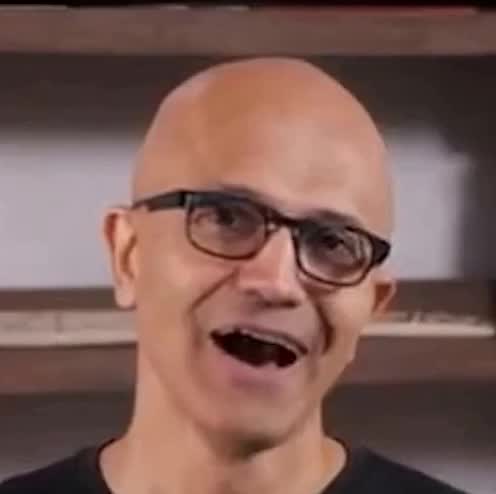}}&
\shortstack{\includegraphics[width=0.33\linewidth]{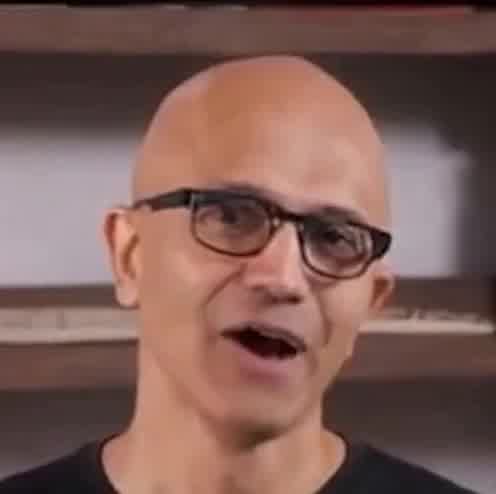}}&
\shortstack{\includegraphics[width=0.33\linewidth]{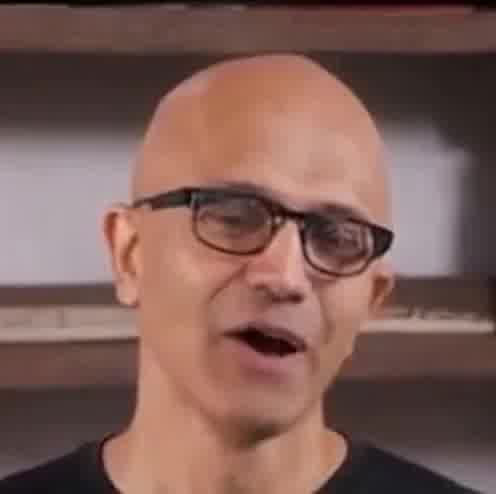}}&
\shortstack{\includegraphics[width=0.33\linewidth]{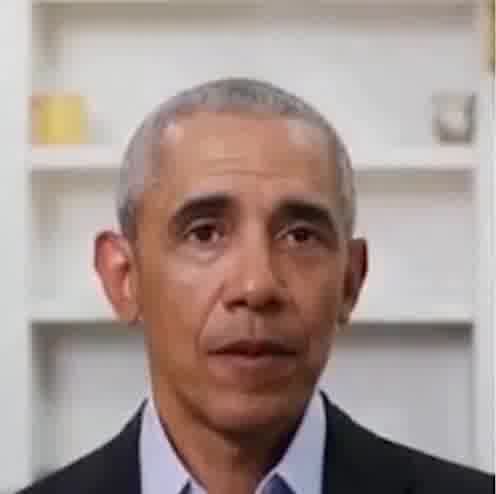}}&
\shortstack{\includegraphics[width=0.33\linewidth]{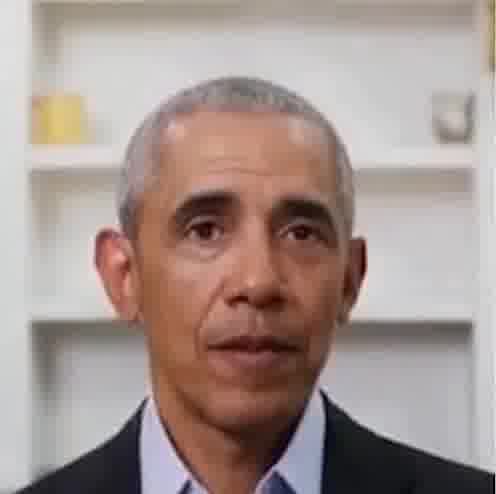}}&
\shortstack{\includegraphics[width=0.33\linewidth]{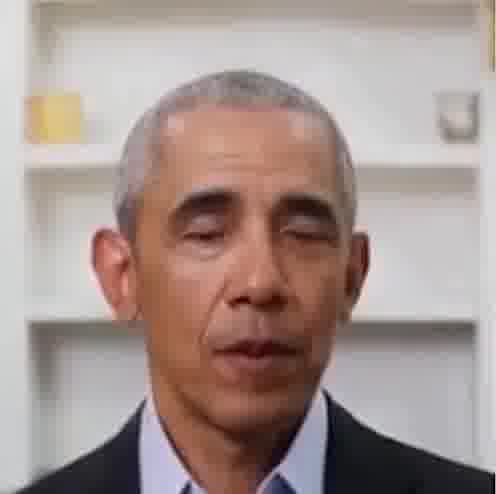}}\\[1pt]
\rotatebox[origin=l]{90}{\hspace{0.8cm} \textbf{[Lu et. al.]}} &
\shortstack{\includegraphics[width=0.33\linewidth]{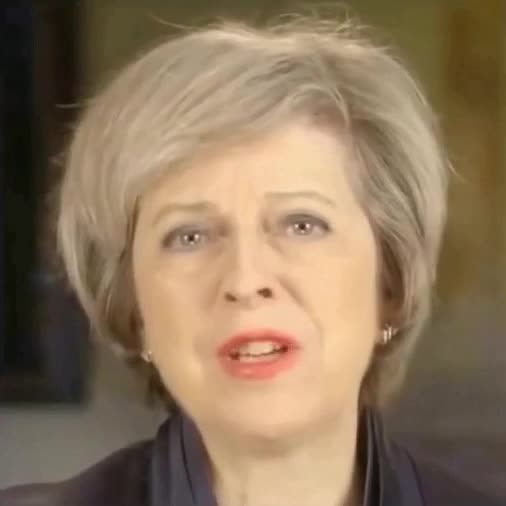}}&
\shortstack{\includegraphics[width=0.33\linewidth]{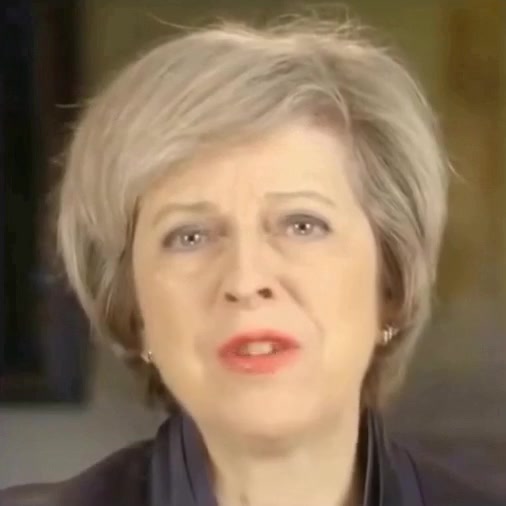}}&
\shortstack{\includegraphics[width=0.33\linewidth]{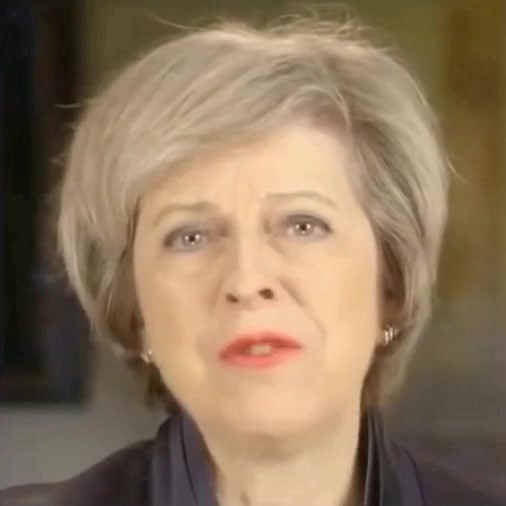}}&
\shortstack{\includegraphics[width=0.33\linewidth]{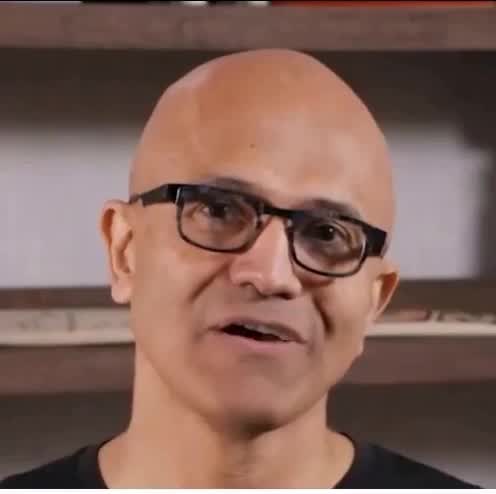}}&
\shortstack{\includegraphics[width=0.33\linewidth]{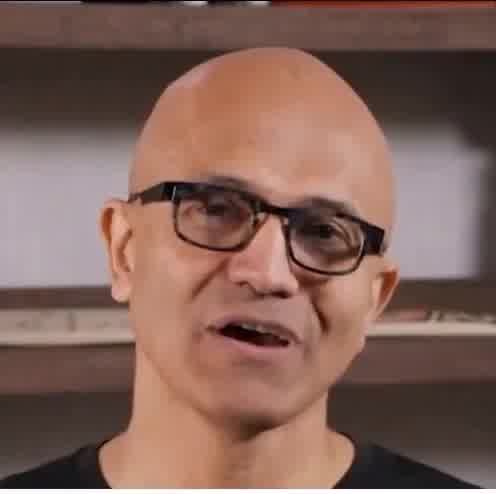}}&
\shortstack{\includegraphics[width=0.33\linewidth]{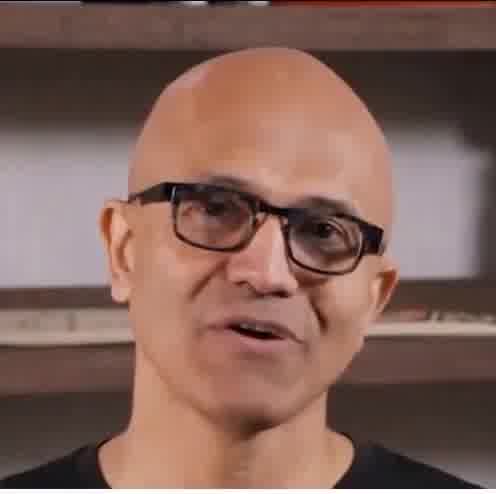}}&
\shortstack{\includegraphics[width=0.33\linewidth]{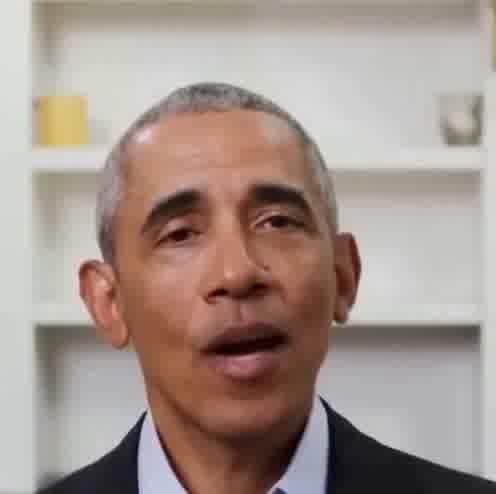}}&
\shortstack{\includegraphics[width=0.33\linewidth]{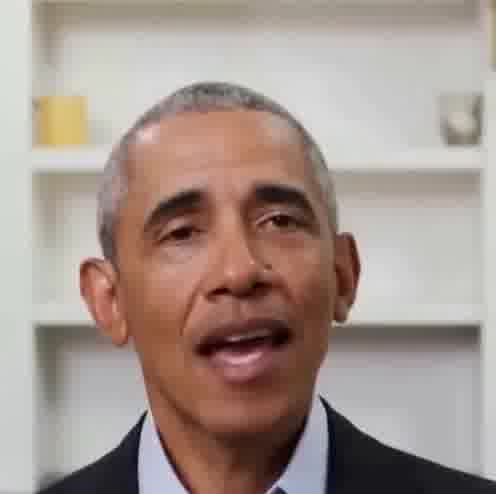}}&
\shortstack{\includegraphics[width=0.33\linewidth]{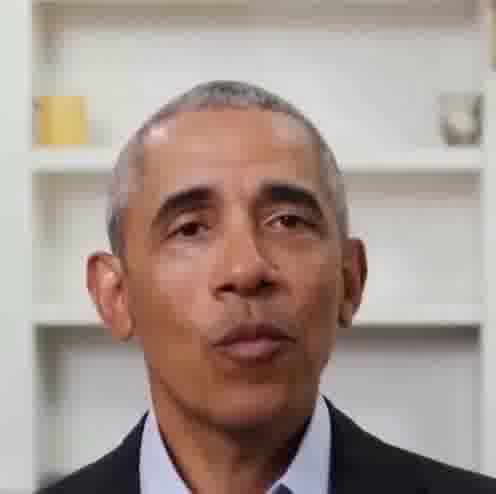}}\\[15pt]
\hline\\
\rotatebox[origin=l]{90}{\hspace{0cm} \textbf{Ours (Original)}} &
\shortstack{\includegraphics[width=0.33\linewidth]{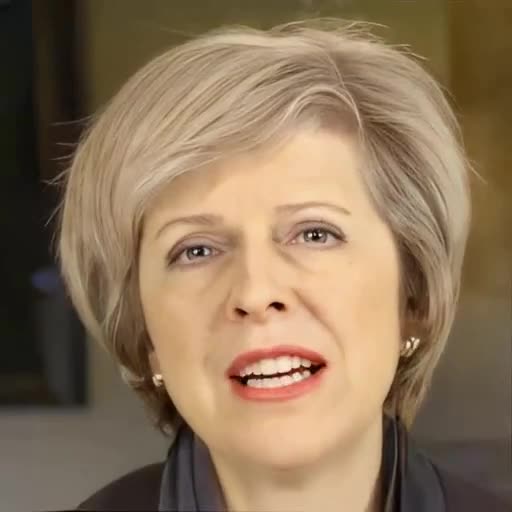}}&
\shortstack{\includegraphics[width=0.33\linewidth]{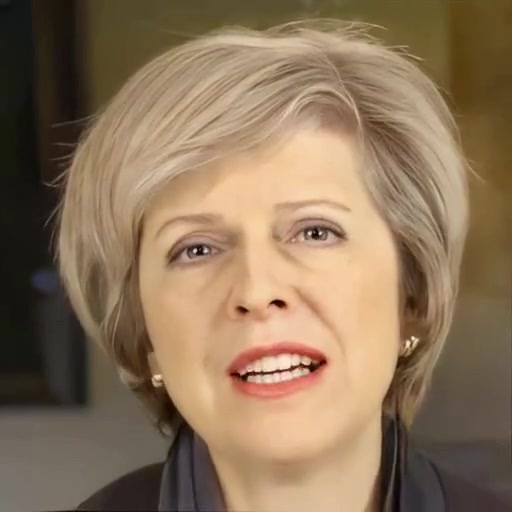}}&
\shortstack{\includegraphics[width=0.33\linewidth]{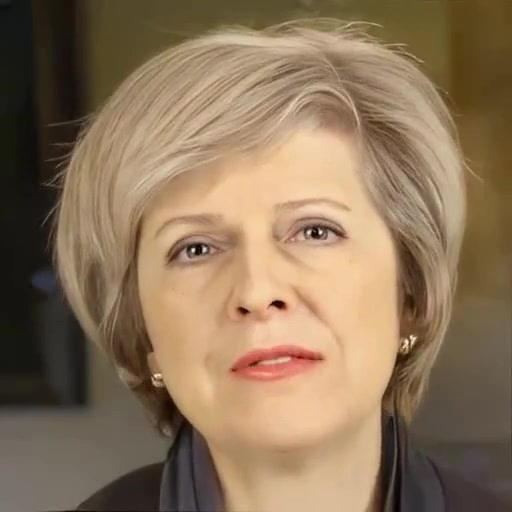}}&
\shortstack{\includegraphics[width=0.33\linewidth]{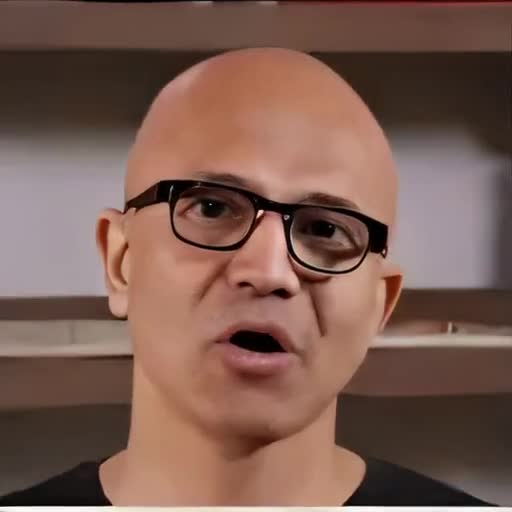}}&
\shortstack{\includegraphics[width=0.33\linewidth]{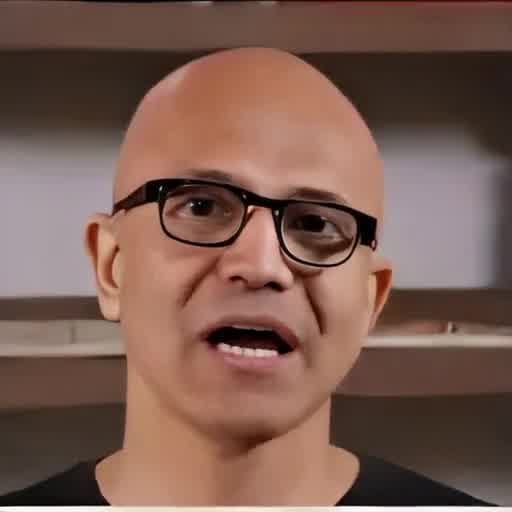}}&
\shortstack{\includegraphics[width=0.33\linewidth]{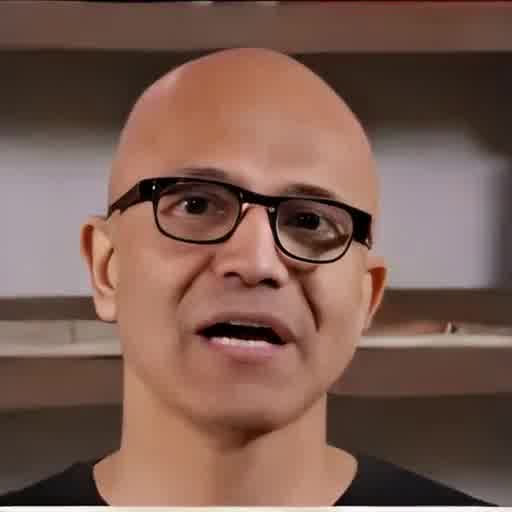}}&
\shortstack{\includegraphics[width=0.33\linewidth]{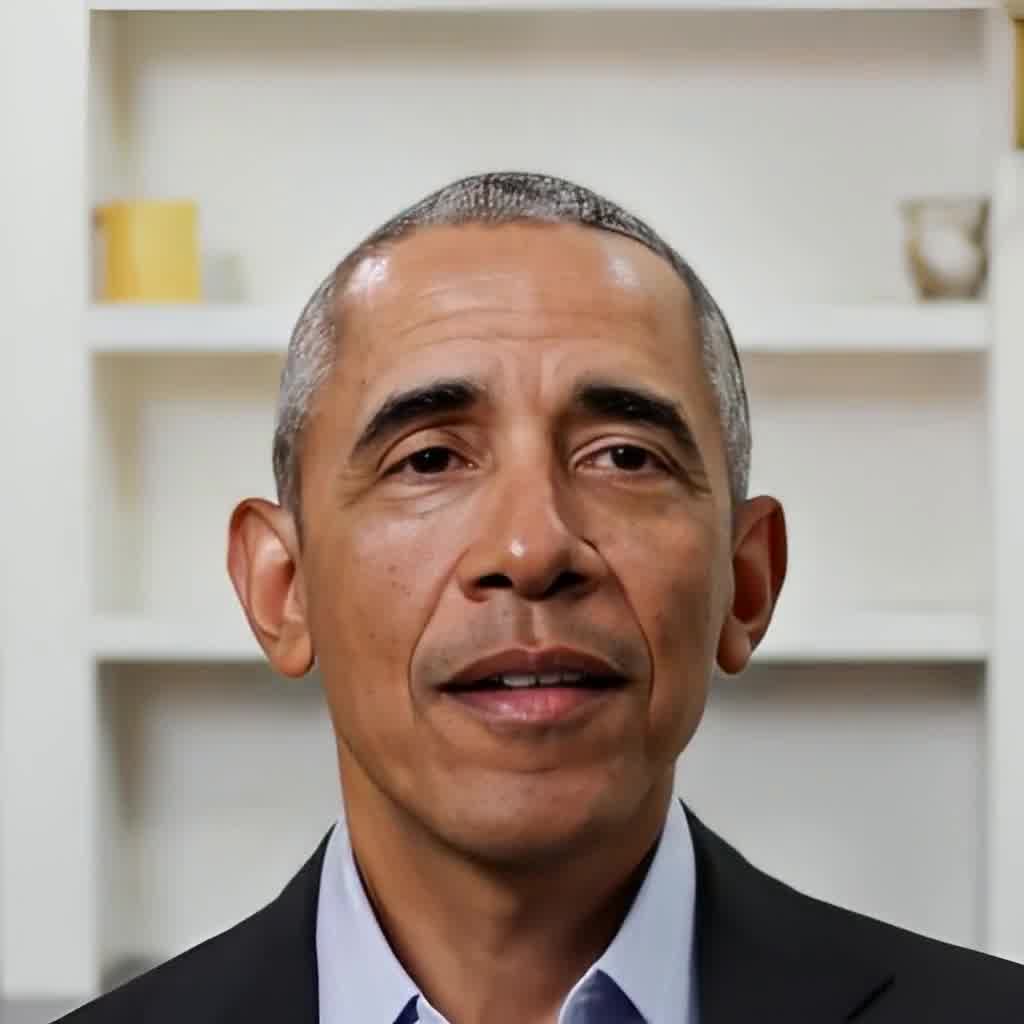}}&
\shortstack{\includegraphics[width=0.33\linewidth]{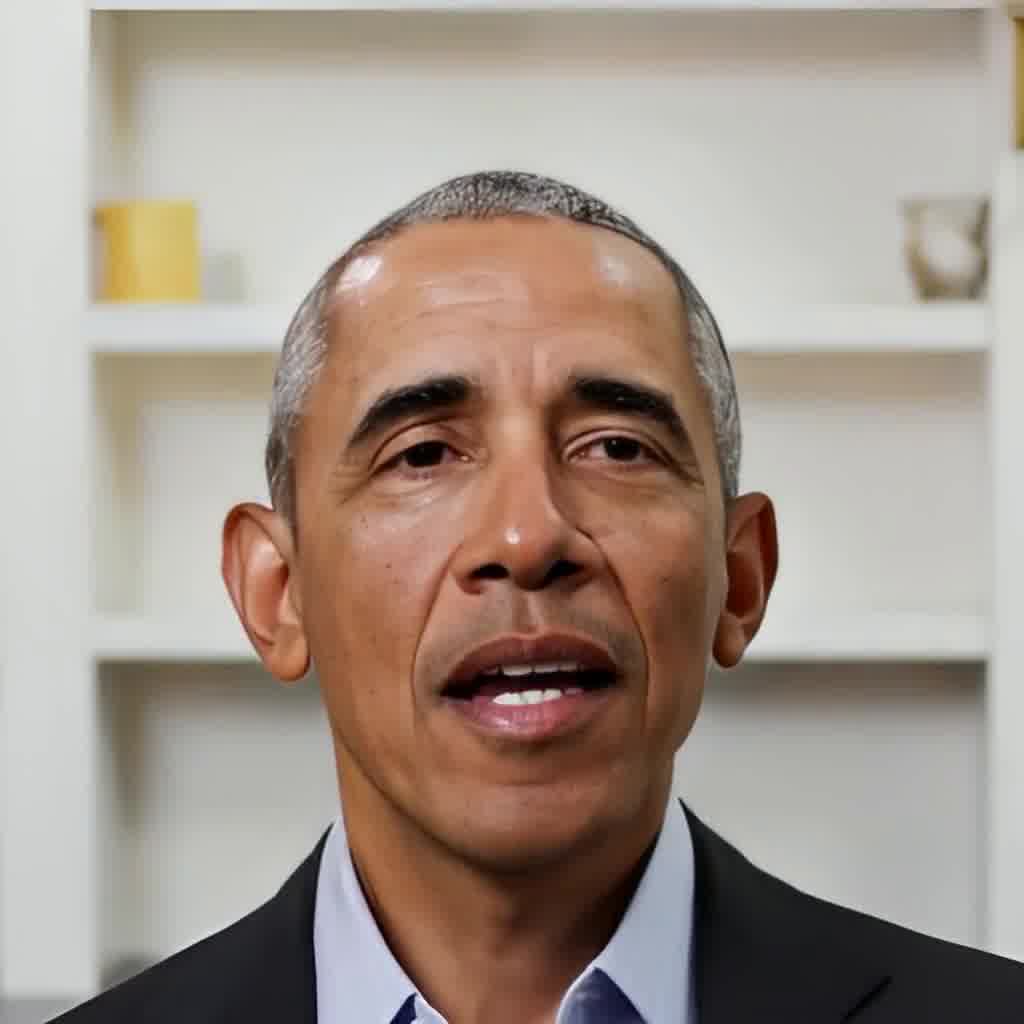}}&
\shortstack{\includegraphics[width=0.33\linewidth]{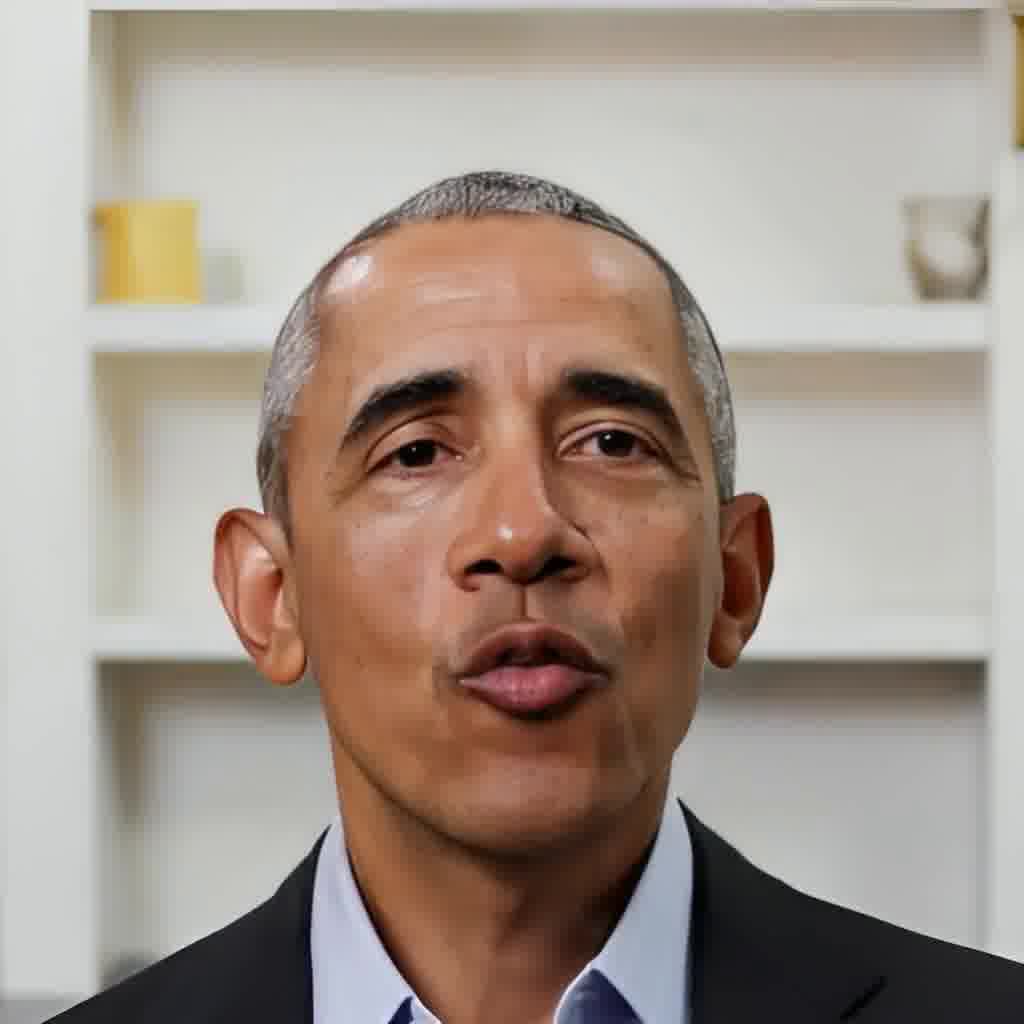}}\\[1pt]
\rotatebox[origin=l]{90}{\hspace{0cm} \textbf{Ours (Cropped)}} &
\shortstack{\includegraphics[width=0.33\linewidth]{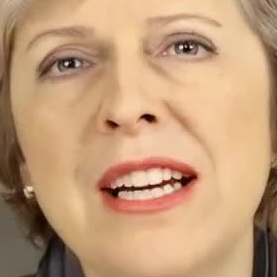}}&
\shortstack{\includegraphics[width=0.33\linewidth]{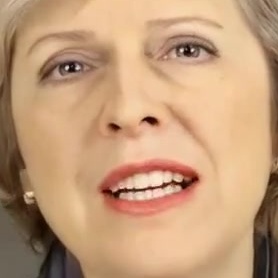}}&
\shortstack{\includegraphics[width=0.33\linewidth]{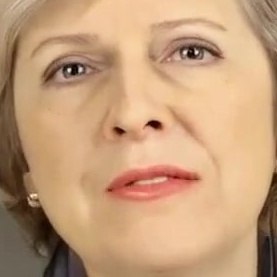}}&
\shortstack{\includegraphics[width=0.33\linewidth]{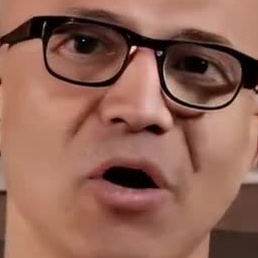}}&
\shortstack{\includegraphics[width=0.33\linewidth]{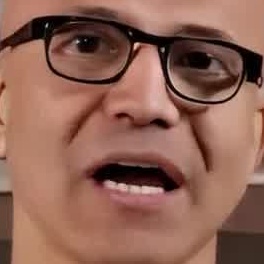}}&
\shortstack{\includegraphics[width=0.33\linewidth]{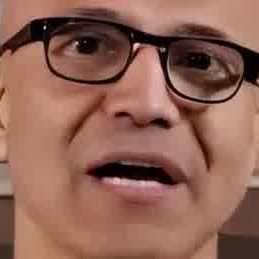}}&
\shortstack{\includegraphics[width=0.33\linewidth]{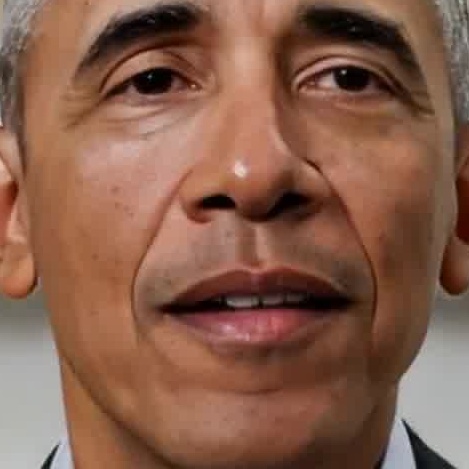}}&
\shortstack{\includegraphics[width=0.33\linewidth]{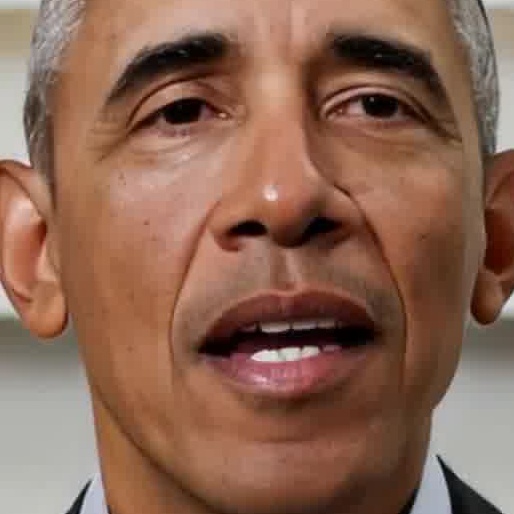}}&
\shortstack{\includegraphics[width=0.33\linewidth]{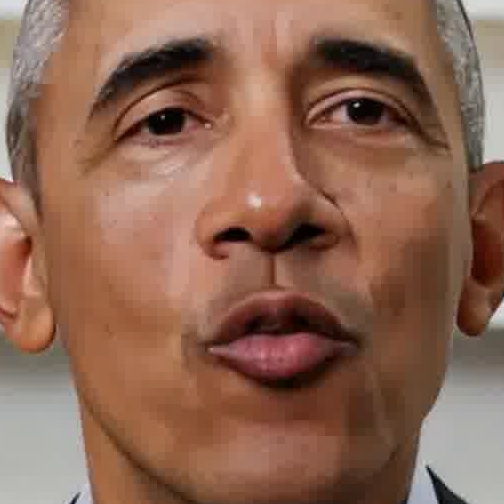}}\\[1pt]
\end{tabular}
}
    \caption{Comparison between different 2D talking head galleries.}
    \label{fig:2DCompare}
\end{figure*}
Table~\ref{tab:ad_method_compare} illustrates the overall capacity comparison between our method and the recent state of the arts. Our proposed method can output a 2D photo-realistic talking head, learn personalized style information, and transfer the specific character style by using only a single input image and an audio stream. In practice, our method can provide high-resolution 2D output and smooth motion based on the audio input. Furthermore, our transfer learning step allows us to quickly adapt and generate new motion based on a specific reference style.

\subsubsection{2D Talking Head Animation}

We compare our method with other image-based methods for 2D talking head generation. Specifically, we compare with~\cite{vougioukas2019end, chen_photo-realistic_2019, zhou2020makelttalkMIT, lu2021liveLSP}. \cite{vougioukas2019end, chen_photo-realistic_2019, zhou2020makelttalkMIT} train their model for unseen face generation. Note that \cite{vougioukas2019end, chen_photo-realistic_2019} generates the talking head animation only on the cropped faces, which fails to capture the head pose motions.

Figure~\ref{fig:2DCompare} shows the qualitative results driven by the audio input in all methods. Since~\cite{zhou2020makelttalkMIT} warps both the background and the talking head, which can lead to the foreground and the background moving together. Besides, the mouth is twisted, and the synthesized region is blurred. ~\cite{lu2021liveLSP} can synthesize sharper images with higher fidelity. \cite{lu2021liveLSP} also disentangles the head motion and the background. However, while~\cite{vougioukas2019end,chen_photo-realistic_2019,zhou2020makelttalkMIT} and our method only requires a single 2D image to generate the corresponding talking head (one-shot synthesis), in~\cite{lu2021liveLSP} work, the authors require a character's video to learn and render the 2D head. This characteristic of~\cite{lu2021liveLSP} shows limitations in practice when it is challenging to collect video data for each target character. Besides, due to the training process, the target talking style mentioned in~\cite{lu2021liveLSP} tends to fit into the visual information of the target character and cannot be transferred to a different target. Finally, although~\cite{lu2021liveLSP} can synthesize sharp images with high quality, the sharpness of teeth and wrinkles are limited in some extreme cases where the mouth and head variations are high.

Compared to these baselines, our method can generate smooth and natural motion for the 2D talking head. We can produce high-resolution realistic photo output while the head foreground and the background are successfully disentangled . Furthermore, our proposed method not only generate realistic and natural motions for talking motions but also for singing styles such as \texttt{ballad}, \texttt{rap}, \texttt{opera}, etc. The styles can be transferred into different characters using our style transfer process. It is worth noting that we only need a single input image to create a high-fidelity 2D talking head animations while being robust to different challenging talking or singing styles.

\subsubsection{Style Transfer Results}
  
\begin{figure}[!ht] 
  \centering
\resizebox{\linewidth}{!}{
\setlength{\tabcolsep}{2pt}
\begin{tabular}{cccccc}

\shortstack{\includegraphics[width=0.33\linewidth]{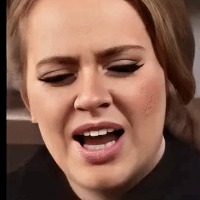}}&
\shortstack{\includegraphics[width=0.33\linewidth]{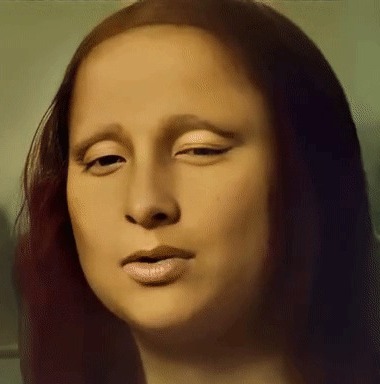}}&
\shortstack{\includegraphics[width=0.33\linewidth]{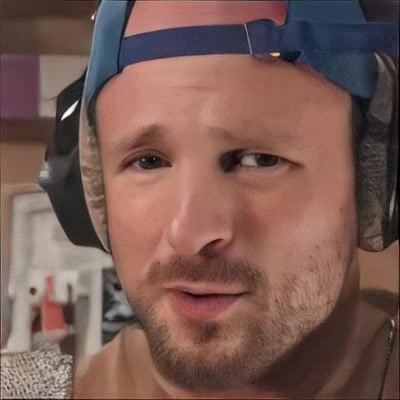}}&
\shortstack{\includegraphics[width=0.33\linewidth]{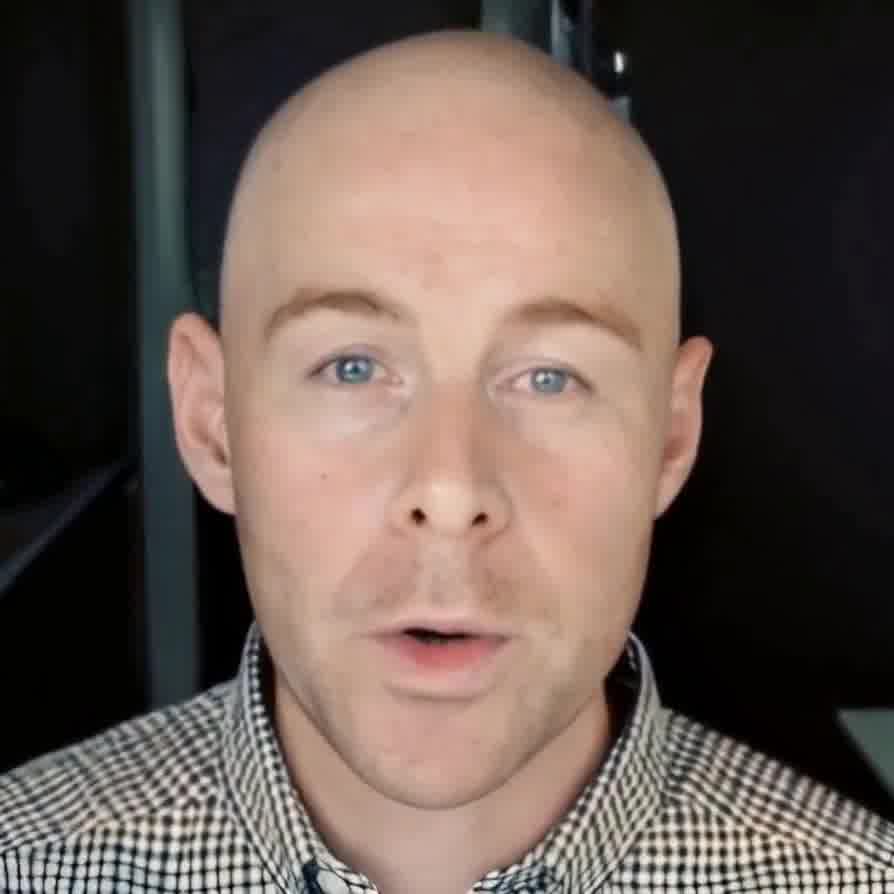}}&
\shortstack{\includegraphics[width=0.33\linewidth]{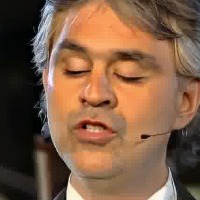}}&
\shortstack{\includegraphics[width=0.33\linewidth]{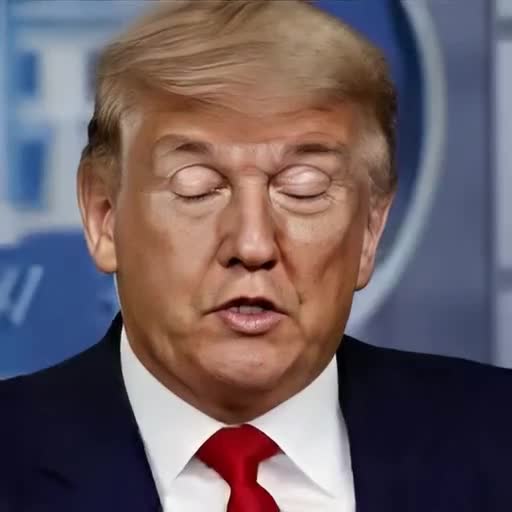}}\\[1pt]
\shortstack{\includegraphics[width=0.33\linewidth]{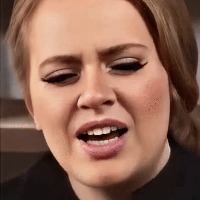}}&
\shortstack{\includegraphics[width=0.33\linewidth]{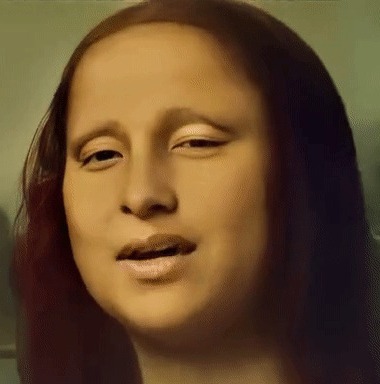}}&
\shortstack{\includegraphics[width=0.33\linewidth]{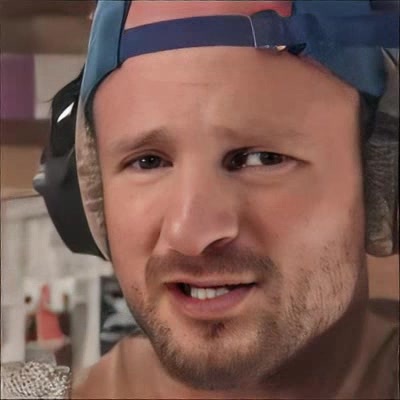}}&
\shortstack{\includegraphics[width=0.33\linewidth]{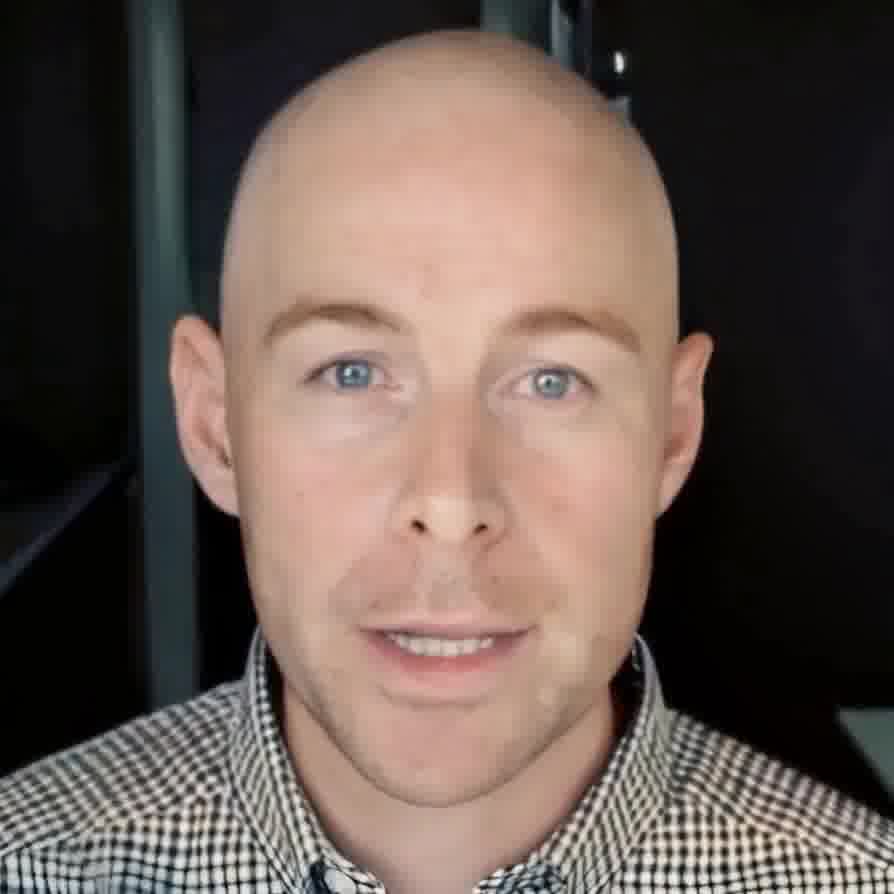}}&
\shortstack{\includegraphics[width=0.33\linewidth]{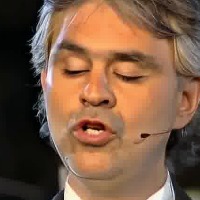}}&
\shortstack{\includegraphics[width=0.33\linewidth]{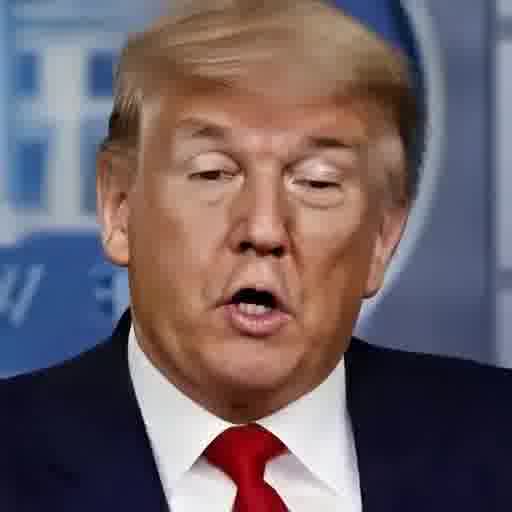}}\\[1pt]
\shortstack{\includegraphics[width=0.33\linewidth]{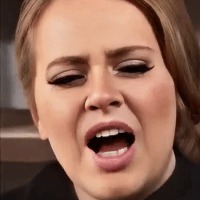} \\  (a)}&
\shortstack{\includegraphics[width=0.33\linewidth]{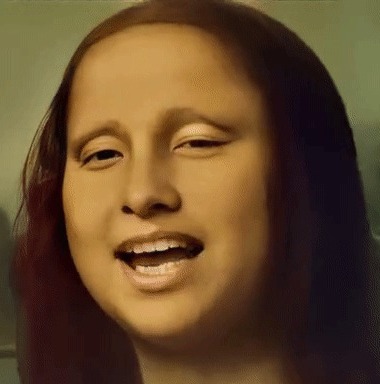} \\ (b)}&
\shortstack{\includegraphics[width=0.33\linewidth]{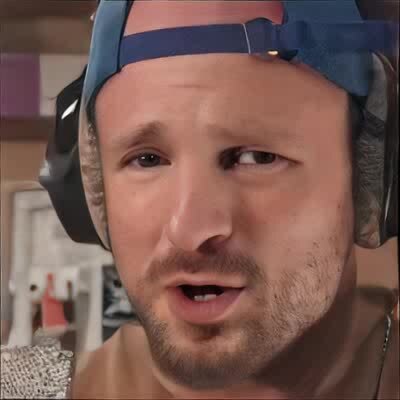} \\ (c)}&
\shortstack{\includegraphics[width=0.33\linewidth]{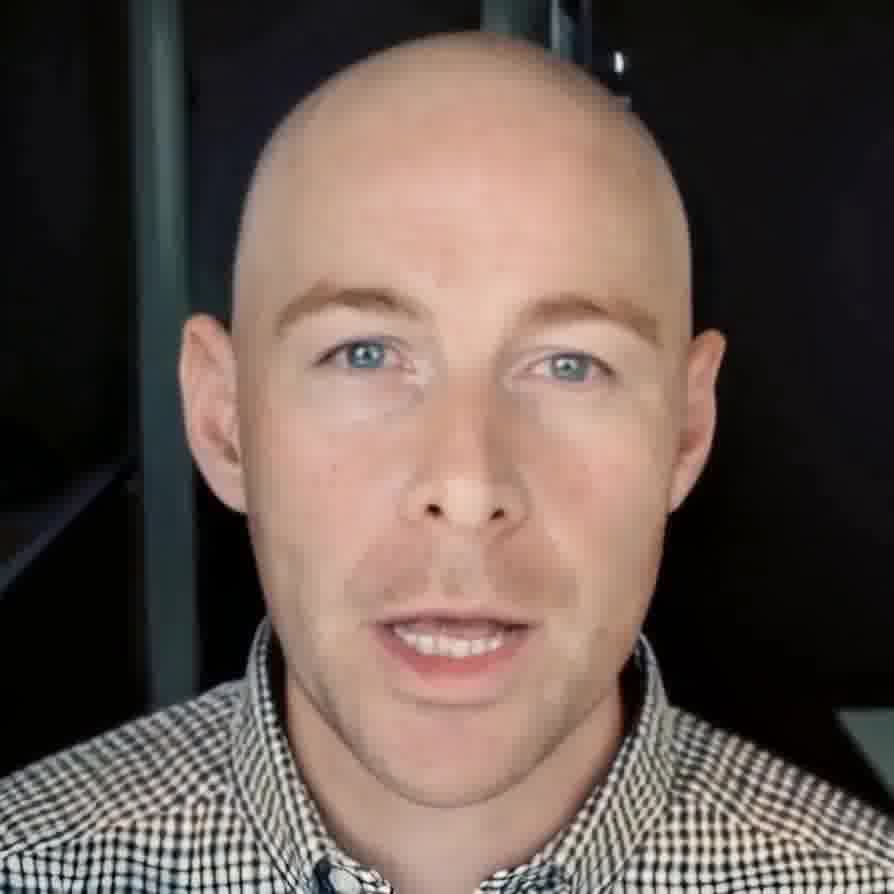} \\ (d)}&
\shortstack{\includegraphics[width=0.33\linewidth]{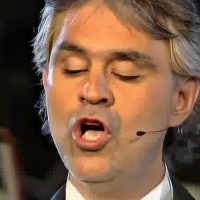} \\ (e)}&
\shortstack{\includegraphics[width=0.33\linewidth]{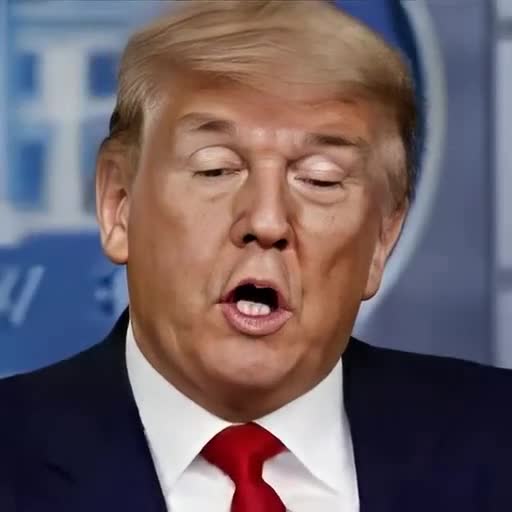} \\ (f)}\\ [1pt]
\end{tabular}
}

    \caption{Our 2D photo-realistic talking head results with different styles. (a), (c), (e) are \texttt{ballad}, \texttt{rap}, and \texttt{opera} style references, respectively; (b), (d), (f) are the corresponding style transfer results. For more details, please visit our demonstration video.}
    \label{fig:styleTransfer}
\vspace{-1.5 pt}

\end{figure}

Figure~\ref{fig:styleTransfer} shows that our method successfully transfers different styles such as \texttt{ballad}, \texttt{rap}, or \texttt{opera} to a new target character. For the \texttt{ballad} style, we use the short singing clip of Adele as the reference. The \texttt{ballad} style usually contains short echoes, a slightly moving head, and closed eyes during the performance.    
For the \texttt{rap} style, the short rapping clip of Mac is used for extracting the \texttt{rap} style reference. \texttt{Rap} style may have rapid head sharking and fast lips movement. 
And for the \texttt{opera} style, we use Andrea's sample clip to obtain the style reference. The \texttt{opera} style has a long echo, a curl of the lips, closed eyes, and slow head movement during the performance.
Note that the audio sequences used for these animations are different from the audio of the style references and unseen during training.

In the \texttt{ballad} case, our method successfully captures the personalized singing style of Adele and transfer it to Mona Lisa's talking head. The initial pose of the Mona Lisa is kept the same as the input static image. 
In the \texttt{rap} style, although the head and mouth motions of the character have extremely active dynamics and high intensity, which is common in fast-rap music, our method is still successful in capturing these behaviors of the animations. 
In the \texttt{opera} case, our method can identify the special mouth and teeth which are unique in \texttt{opera} style.
We note that our method can learn and transfer style to animate any arbitrary image. Besides, our proposed method is not only able to animate real person but also non-realistic ones such as human-like portraits, arts, or painting images (e.g., Mona Lisa). For more details, please refer to the demonstration video.

\begin{figure}[!ht] 
 \centering

\resizebox{\linewidth}{!}{
\setlength{\tabcolsep}{3pt}
\begin{tabular}{cccccc}
\rotatebox[origin=l]{90}{\hspace{0.3cm}  \textbf{Zhou \etal \cite{zhou2020makelttalkMIT}}} &
\shortstack{\includegraphics[width=0.33\linewidth]{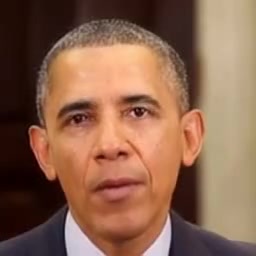}}&
\shortstack{\includegraphics[width=0.33\linewidth]{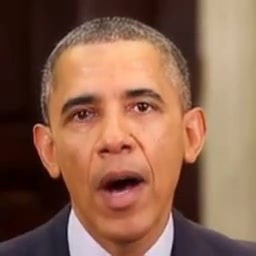}}&
\shortstack{\includegraphics[width=0.33\linewidth]{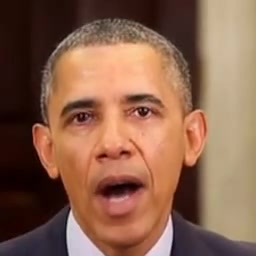}}&
\shortstack{\includegraphics[width=0.33\linewidth]{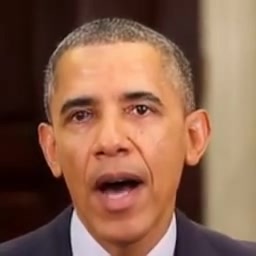}}\\[1pt]
\rotatebox[origin=l]{90}{\hspace{0.2cm} \textbf{Lu \etal\cite{lu2021liveLSP}}} &
\shortstack{\includegraphics[width=0.33\linewidth]{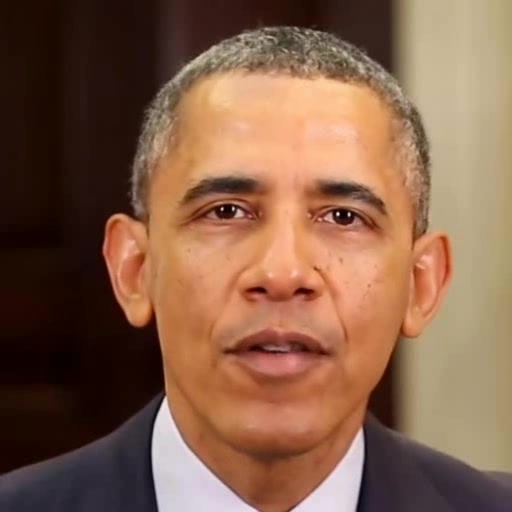}}&
\shortstack{\includegraphics[width=0.33\linewidth]{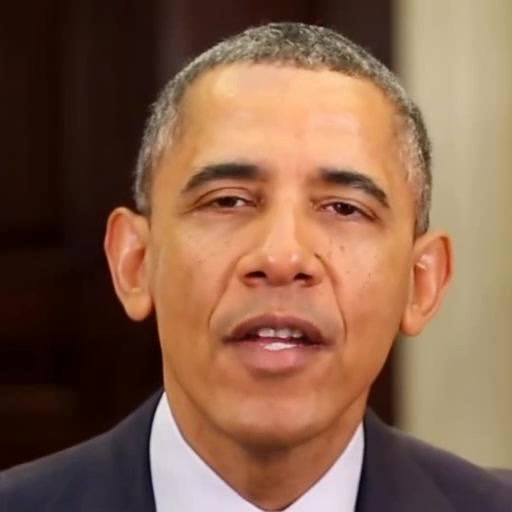}}&
\shortstack{\includegraphics[width=0.33\linewidth]{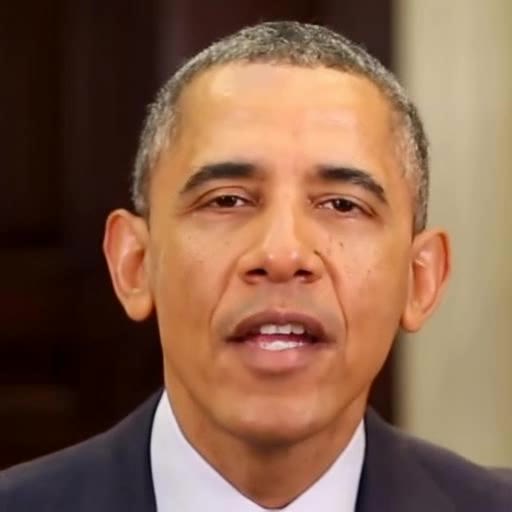}}&
\shortstack{\includegraphics[width=0.33\linewidth]{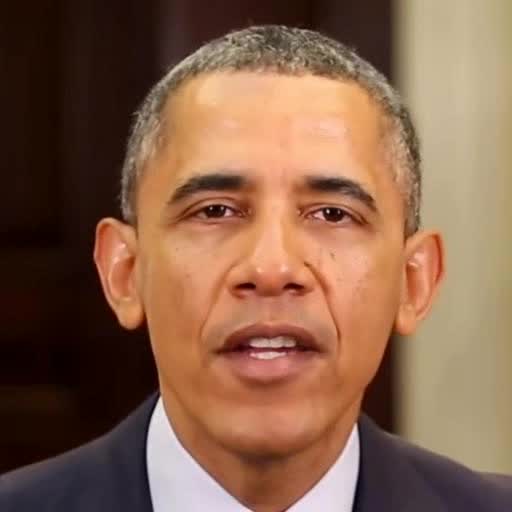}}\\[1pt]
\rotatebox[origin=l]{90}{\hspace{0.8cm} \textbf{Ours}} &
\shortstack{\includegraphics[width=0.33\linewidth]{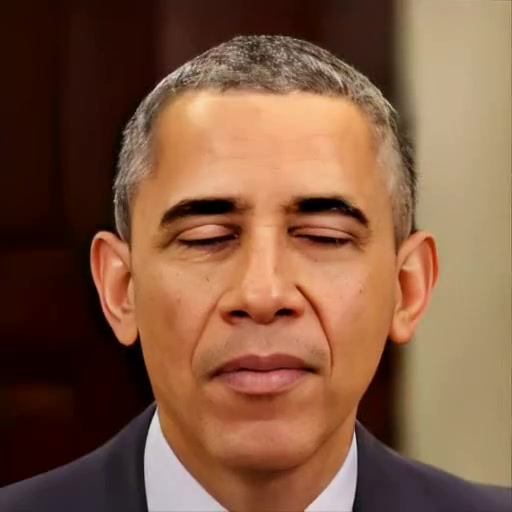}}&
\shortstack{\includegraphics[width=0.33\linewidth]{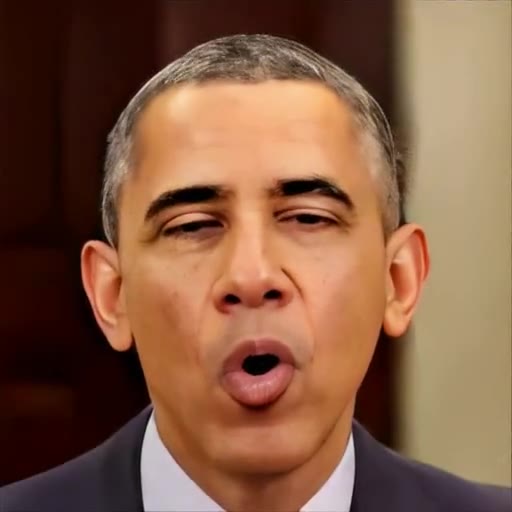}}&
\shortstack{\includegraphics[width=0.33\linewidth]{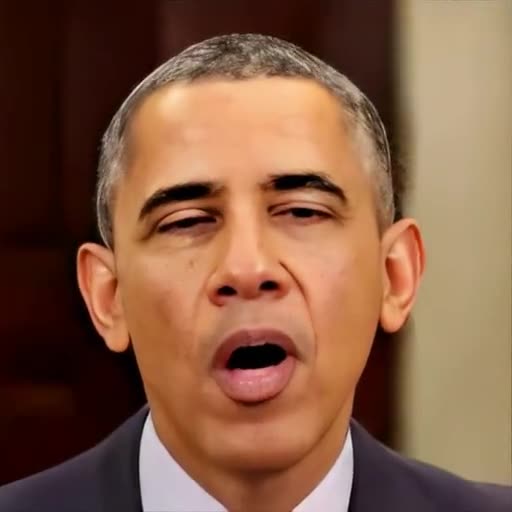}}&
\shortstack{\includegraphics[width=0.33\linewidth]{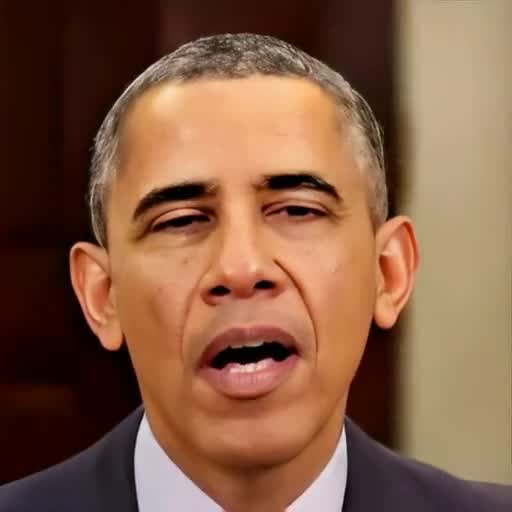}}\\[1pt]

\vspace{-5ex}
\rotatebox[origin=l]{90}{\hspace{0.4cm} \textbf{Lyric}} &
\shortstack{\includegraphics[width=0.2\linewidth]{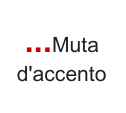}}&
\shortstack{\includegraphics[width=0.2\linewidth]{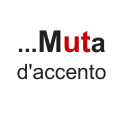}}&
\shortstack{\includegraphics[width=0.2\linewidth]{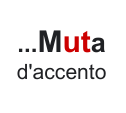}}&
\shortstack{\includegraphics[width=0.2\linewidth]{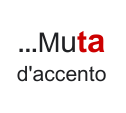}}\\[1pt]

\end{tabular}
}
\vspace{0ex}
    \caption{Comparison between different 2D talking head galleries on \texttt{opera} style. Our method generates more natural and realistic motion, especially around the mouth and the eye of the character.}
    \label{fig:2DStyleComparison}

\end{figure}

Figure~\ref{fig:2DStyleComparison} shows the comparison between our method and recent works on 2D photorealistic talking head animation~\cite{zhou2020makelttalkMIT,lu2021liveLSP} when the character sings an opera song. Focusing on the mouth, we notice that our method produces better results in mouth motion variance and eyes expression compared to the results from~\cite{zhou2020makelttalkMIT} and ~\cite{lu2021liveLSP}. Specifically, in~\cite{zhou2020makelttalkMIT}, the visual fidelity of teeth and pores, as well as the realization of the mouth shape and motion, are not well presented. In~\cite{lu2021liveLSP}, although the quality is reasonable, the mouth shape is rigid and does not present well the pose of lips and eyes in the \texttt{opera} style. The results confirm that our proposed method achieves better lip-synchronization in such extreme cases in comparison with other baselines.

\begin{figure}[!ht] 
  \centering
\resizebox{\linewidth}{!}{
\setlength{\tabcolsep}{3pt}
\begin{tabular}{ccccc}
\rotatebox[origin=l]{90}{\hspace{0.3cm} \textbf{Neutral Style}} &
\shortstack{\includegraphics[width=0.33\linewidth]{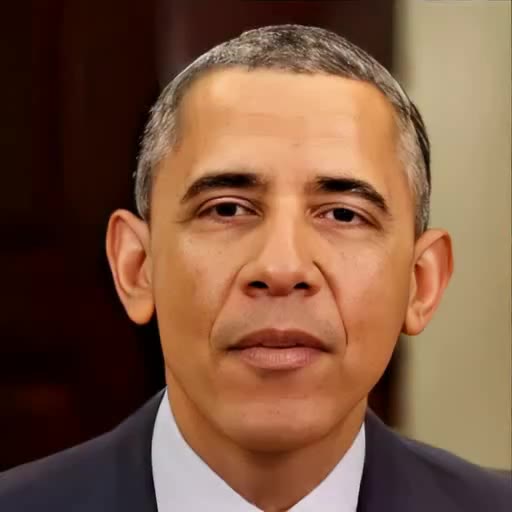}}&
\shortstack{\includegraphics[width=0.33\linewidth]{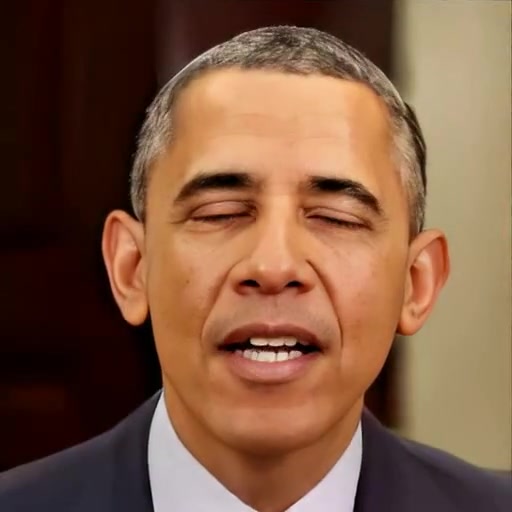}}&
\shortstack{\includegraphics[width=0.33\linewidth]{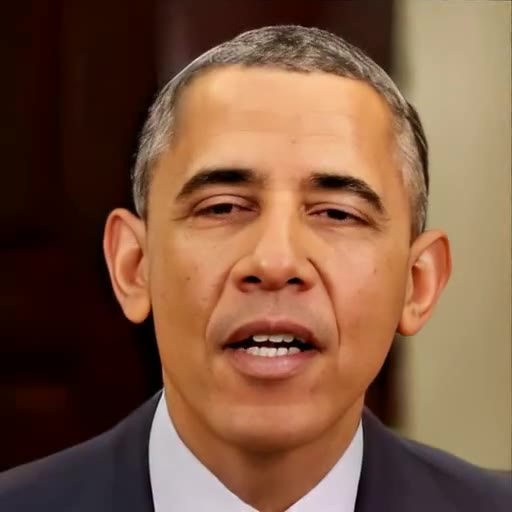}}&
\shortstack{\includegraphics[width=0.33\linewidth]{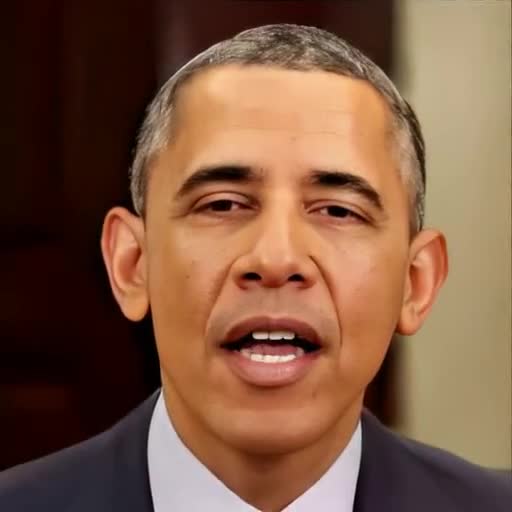}}\\[1pt]
\rotatebox[origin=l]{90}{\hspace{0.2cm}  \textbf{Opera Style}} &
\shortstack{\includegraphics[width=0.33\linewidth]{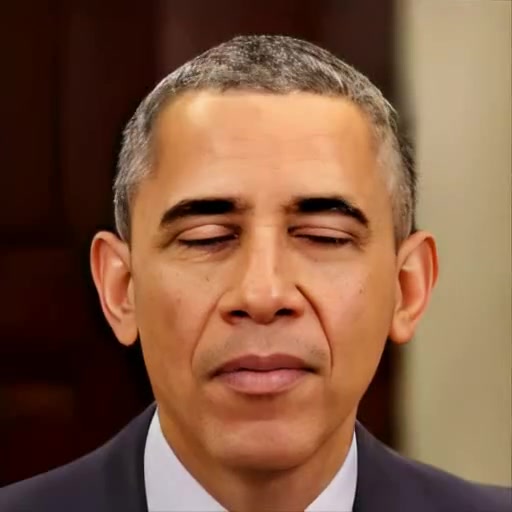}}&
\shortstack{\includegraphics[width=0.33\linewidth]{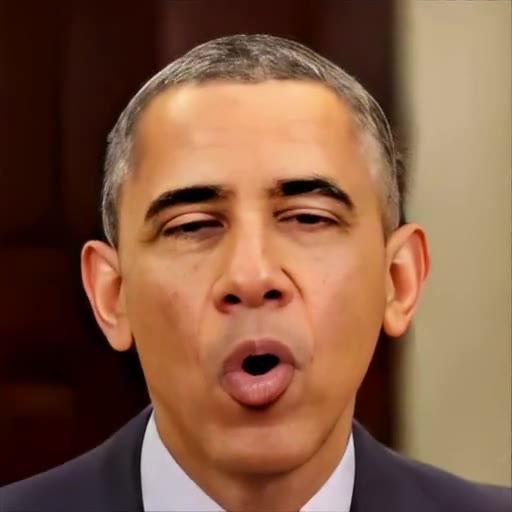}}&
\shortstack{\includegraphics[width=0.33\linewidth]{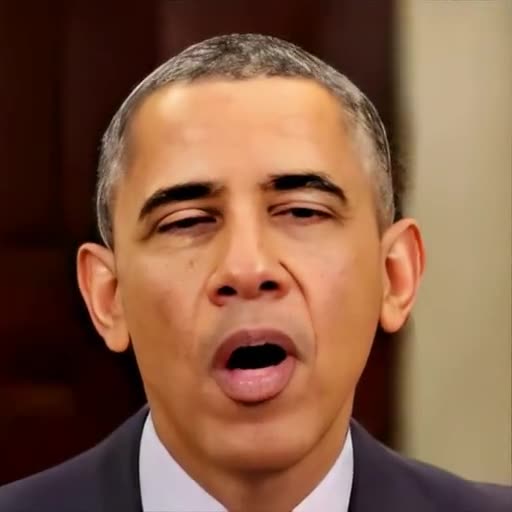}}&
\shortstack{\includegraphics[width=0.33\linewidth]{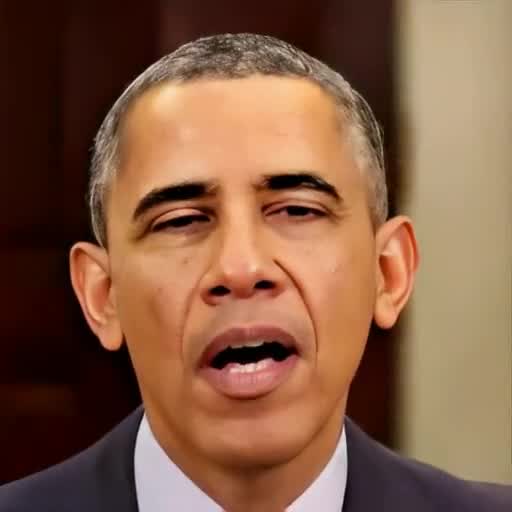}}\\[1pt]
\rotatebox[origin=l]{90}{\hspace{0.3cm} \textbf{Rap Style}} &
\shortstack{\includegraphics[width=0.33\linewidth]{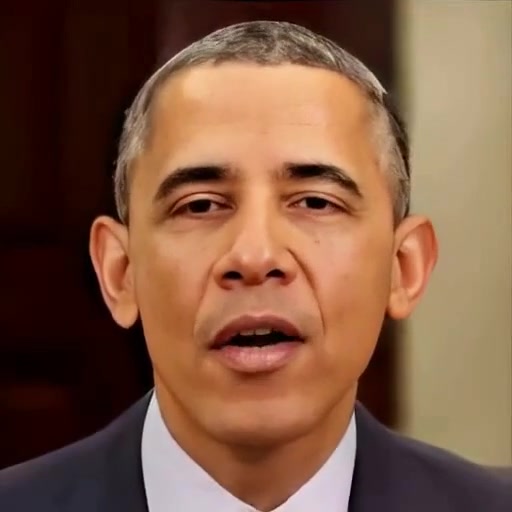}}&
\shortstack{\includegraphics[width=0.33\linewidth]{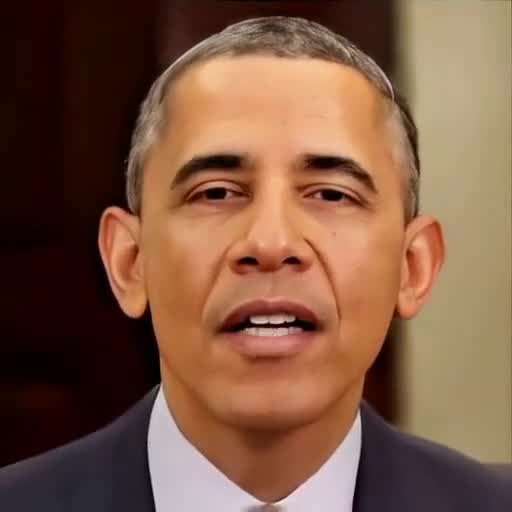}}&
\shortstack{\includegraphics[width=0.33\linewidth]{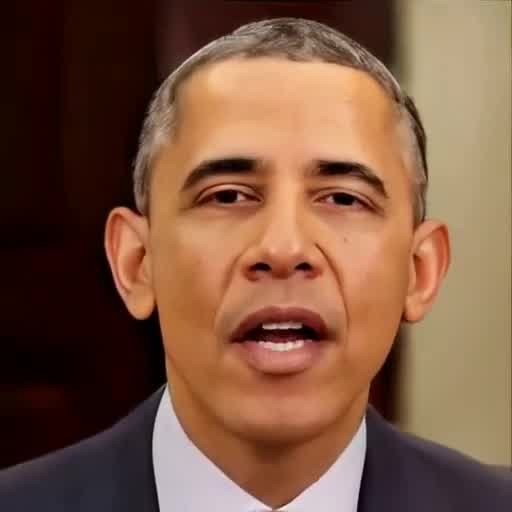}}&
\shortstack{\includegraphics[width=0.33\linewidth]{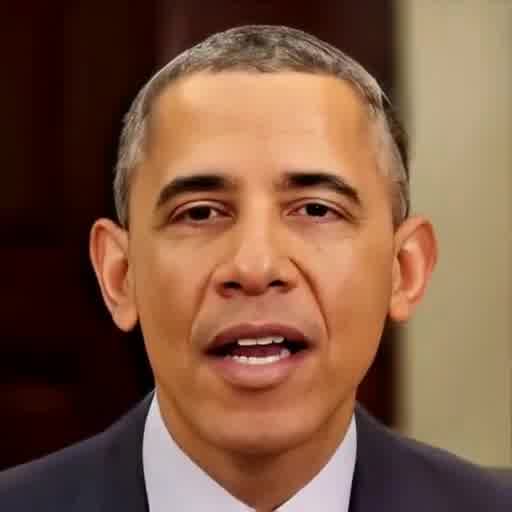}}\\[1pt]
\rotatebox[origin=l]{90}{\hspace{0.2cm} \textbf{Ballad Style}} &
\shortstack{\includegraphics[width=0.33\linewidth]{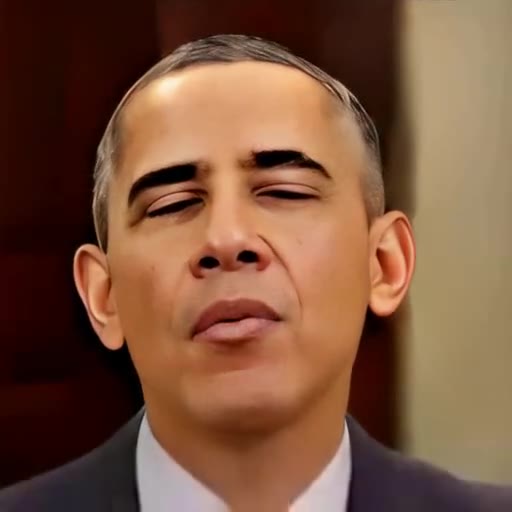}}&
\shortstack{\includegraphics[width=0.33\linewidth]{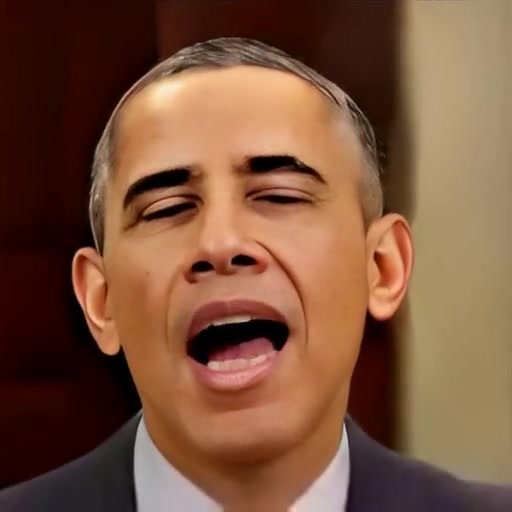}}&
\shortstack{\includegraphics[width=0.33\linewidth]{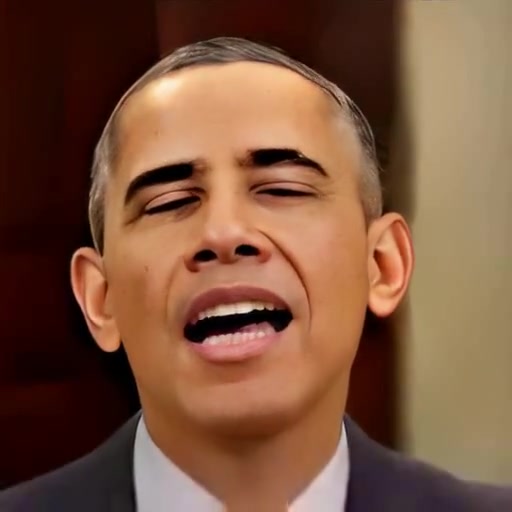}}&
\shortstack{\includegraphics[width=0.33\linewidth]{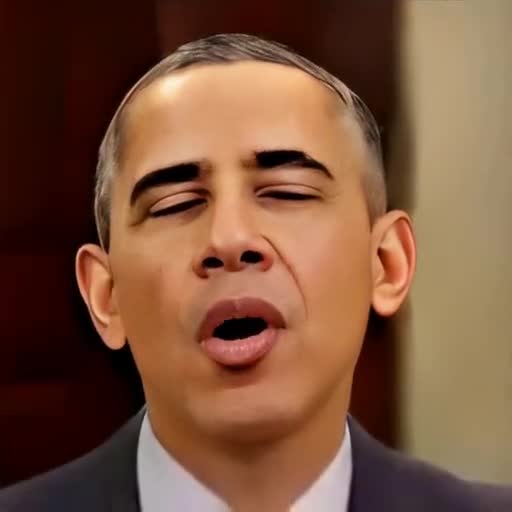}}\\[1pt]
\end{tabular}
}
    \caption{Comparison between different styles in the same input image.}
    \label{fig:fixedImg}
\end{figure}

Figure~\ref{fig:fixedImg} shows the comparison between different styles when they are presented in a fixed input image to generate talking heads. The results illustrate the differences of various styles affecting the 2D talking head animation. Specifically, \texttt{neural} style shows how natural Obama is (in terms of head poses, eye contact, and mouth motions) during his given speech. \texttt{Opera} style focuses on eyes closures, has a curl of the lips, and slow head movement. \texttt{Rap} style encourages the head and mouth motions of the character to be highly dynamic. \texttt{Ballad} style has high variation of mouth motions during the performance, short echoes, slightly moving head, and closed eyes.

\begin{figure}[!ht] 
  \centering
\resizebox{\linewidth}{!}{
\setlength{\tabcolsep}{3pt}
\begin{tabular}{ccccc}
\rotatebox[origin=l]{90}{\hspace{0.3 cm} \textbf{Neutral Style}} &
\shortstack{\includegraphics[width=0.33\linewidth]{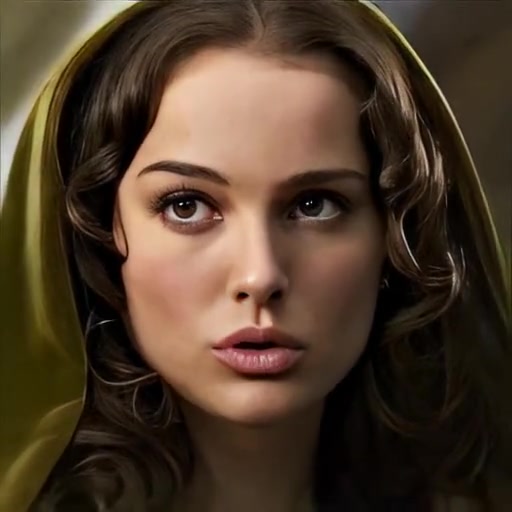}}&
\shortstack{\includegraphics[width=0.33\linewidth]{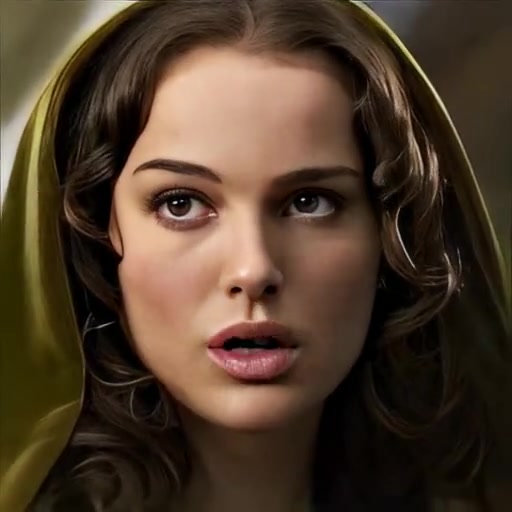}}&
\shortstack{\includegraphics[width=0.33\linewidth]{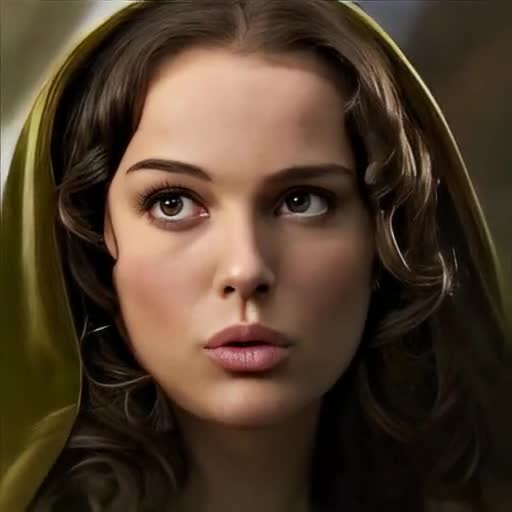}}&
\shortstack{\includegraphics[width=0.33\linewidth]{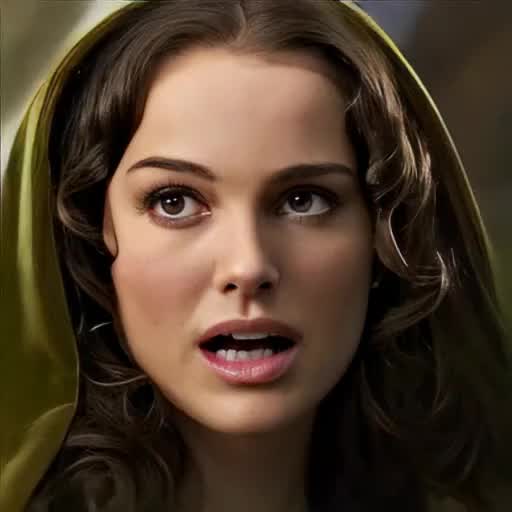}}\\[1pt]
\rotatebox[origin=l]{90}{\hspace{0.3cm}  \textbf{Opera Style}} &
\shortstack{\includegraphics[width=0.33\linewidth]{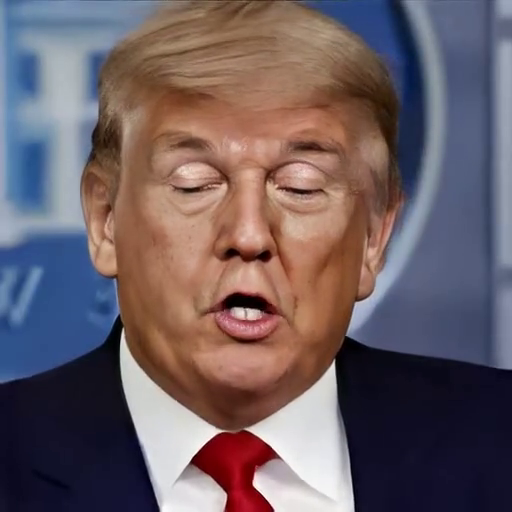}}&
\shortstack{\includegraphics[width=0.33\linewidth]{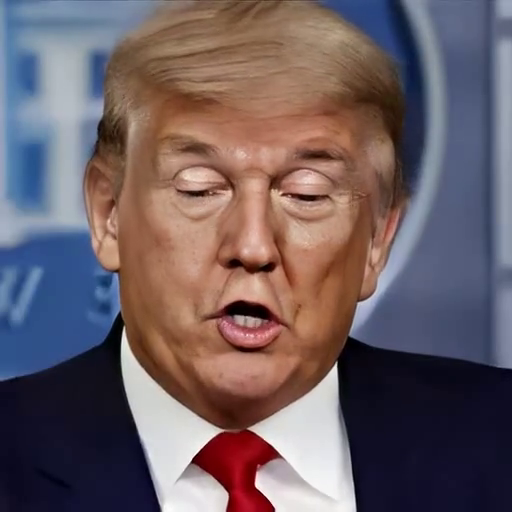}}&
\shortstack{\includegraphics[width=0.33\linewidth]{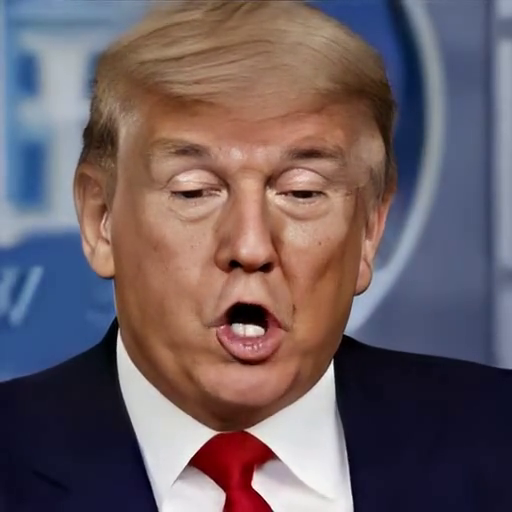}}&
\shortstack{\includegraphics[width=0.33\linewidth]{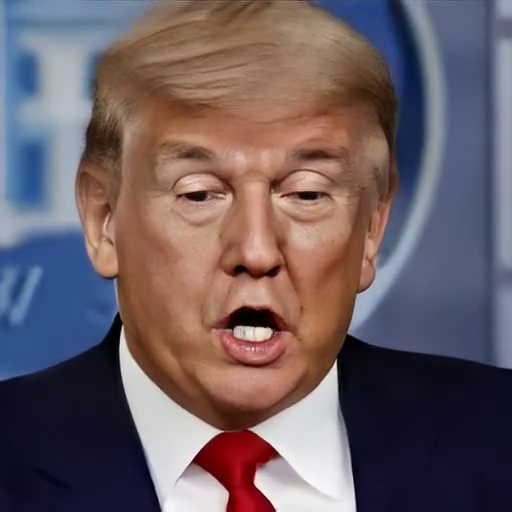}}\\[1pt]
\rotatebox[origin=l]{90}{\hspace{0.6cm} \textbf{Rap Style}} &
\shortstack{\includegraphics[width=0.33\linewidth]{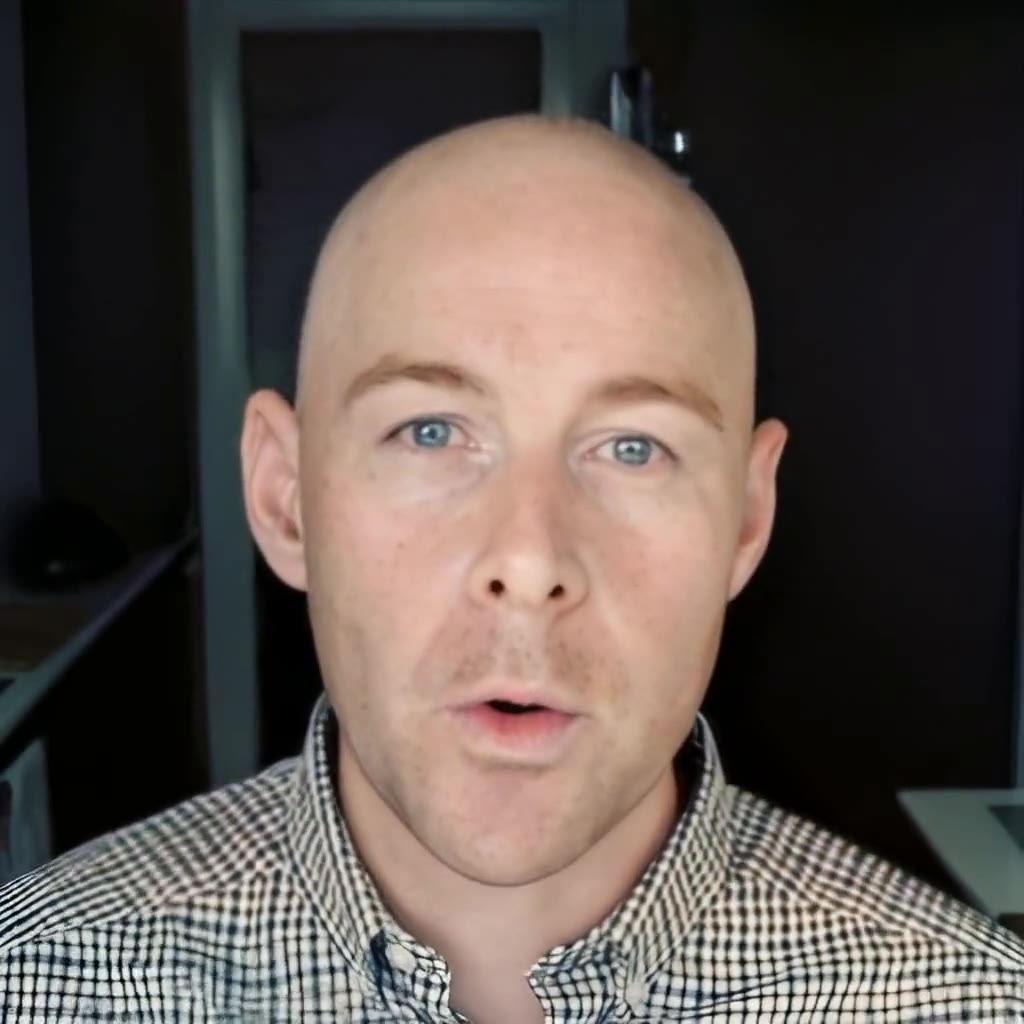}}&
\shortstack{\includegraphics[width=0.33\linewidth]{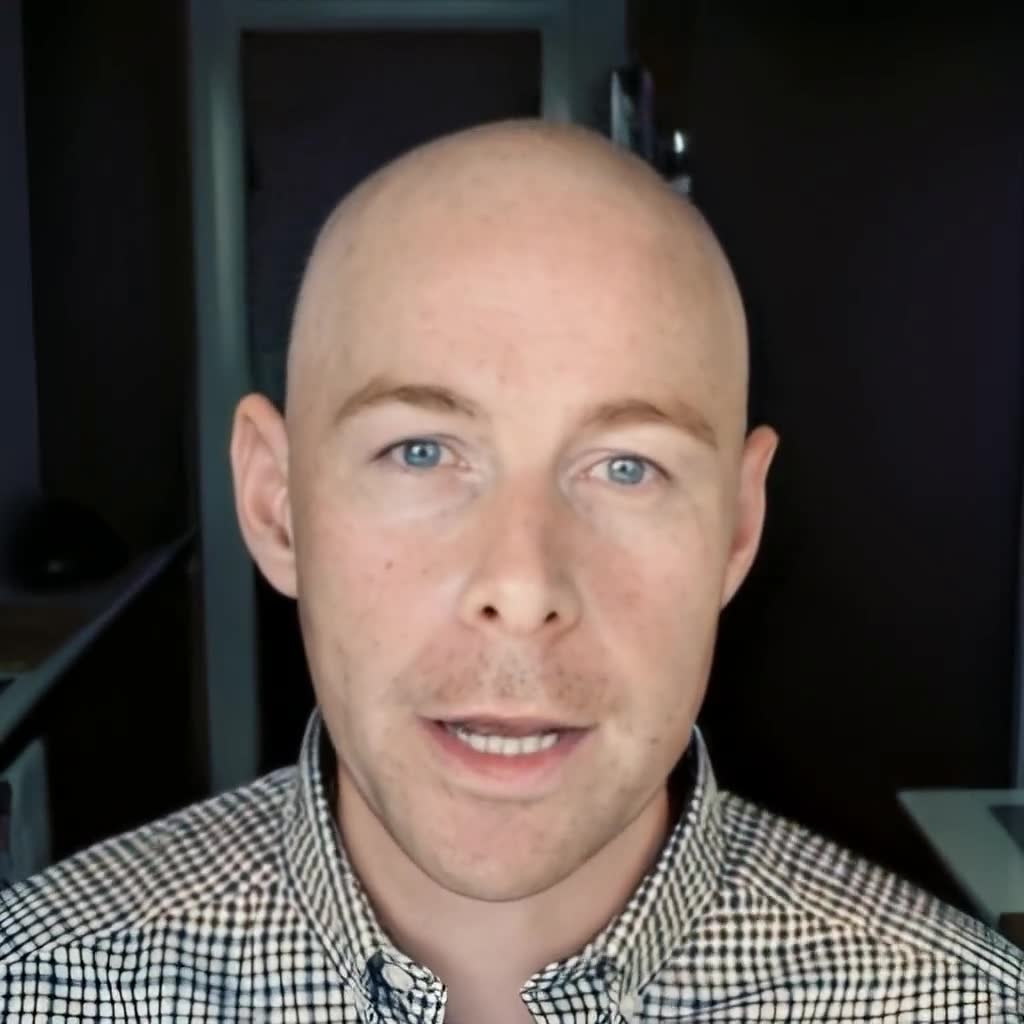}}&
\shortstack{\includegraphics[width=0.33\linewidth]{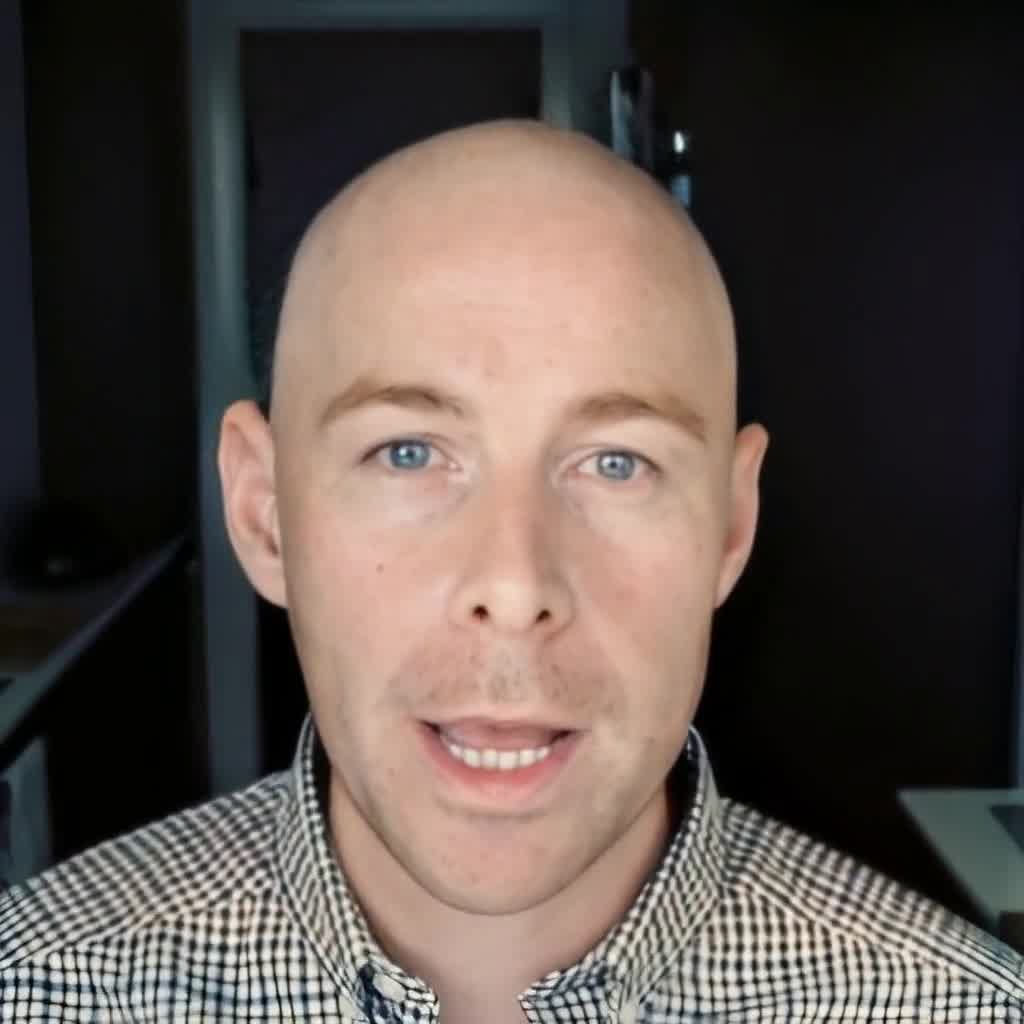}}&
\shortstack{\includegraphics[width=0.33\linewidth]{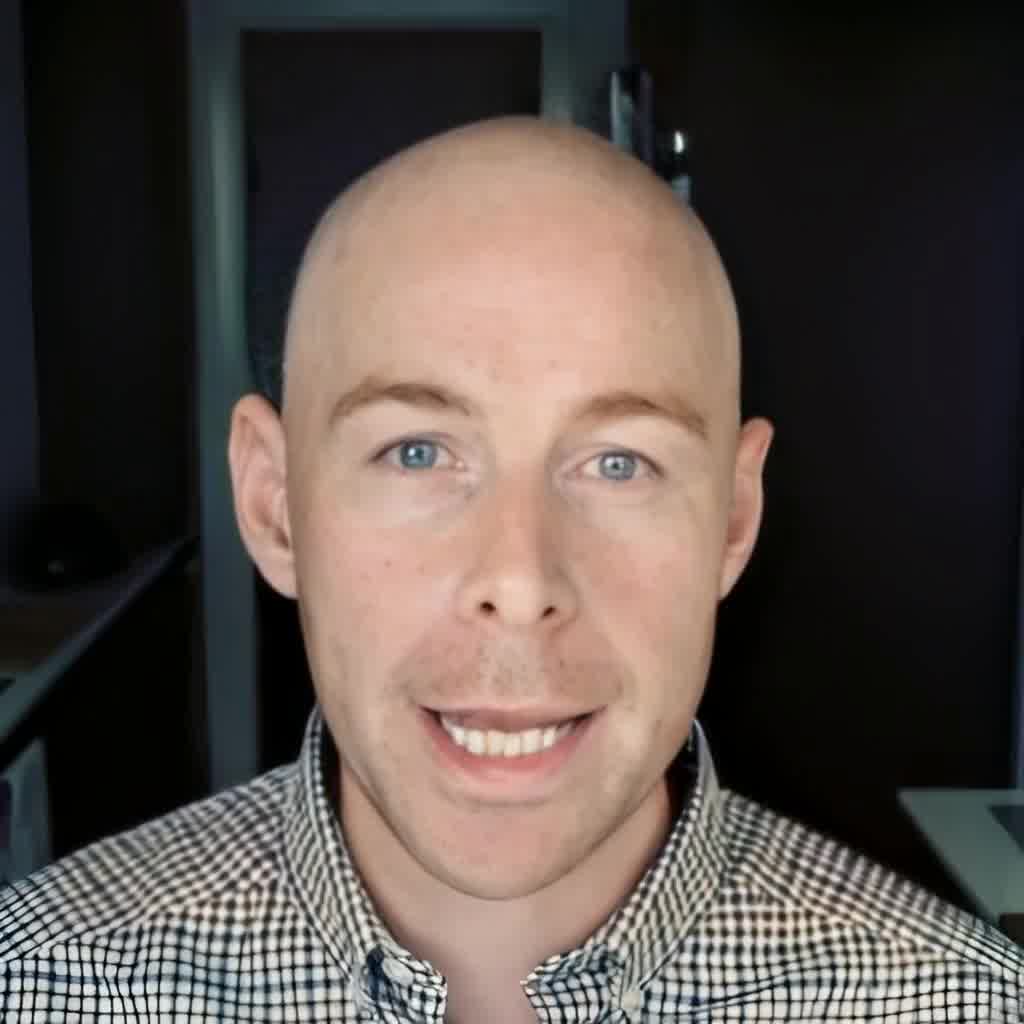}}\\[1pt]
\rotatebox[origin=l]{90}{\hspace{0.4cm} \textbf{Ballad Style}} &
\shortstack{\includegraphics[width=0.33\linewidth]{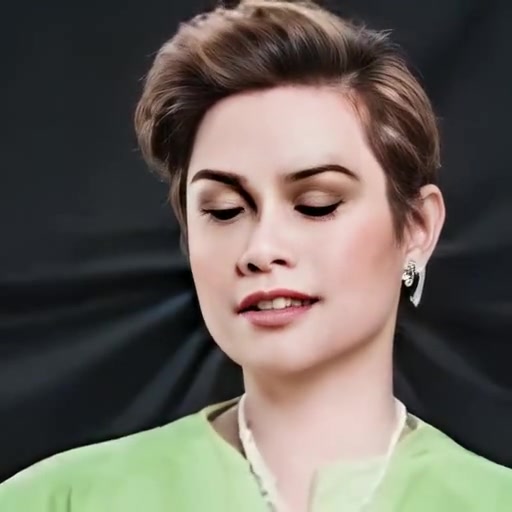}}&
\shortstack{\includegraphics[width=0.33\linewidth]{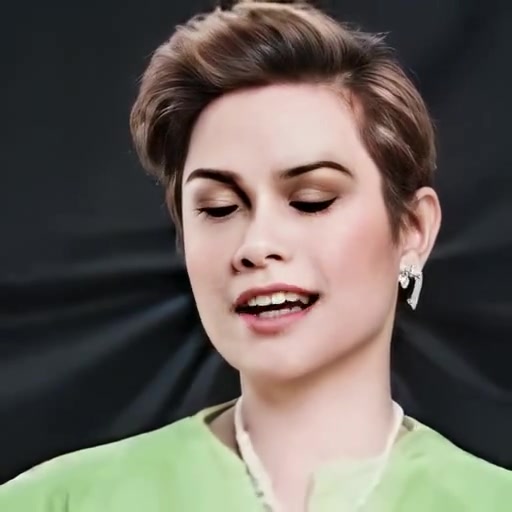}}&
\shortstack{\includegraphics[width=0.33\linewidth]{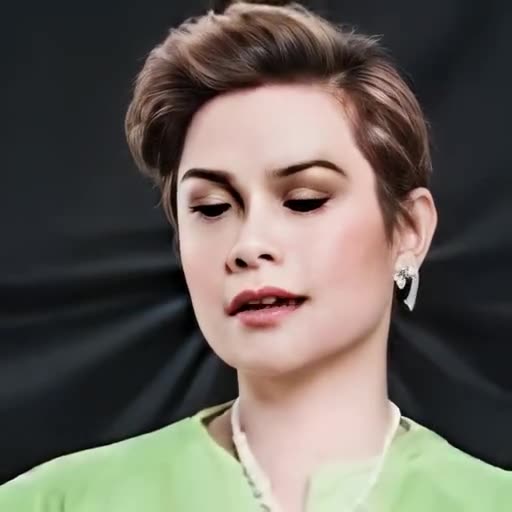}}&
\shortstack{\includegraphics[width=0.33\linewidth]{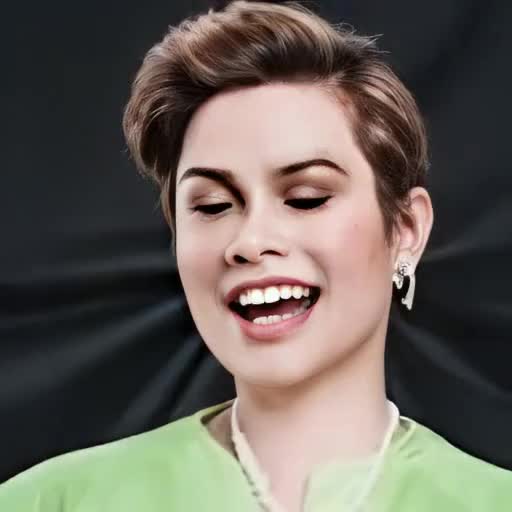}}\\[1pt]
\rotatebox[origin=l]{90}{\hspace{0.2cm} \textbf{Lyric}} &
\shortstack{\includegraphics[width=0.15\linewidth]{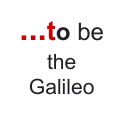}}&
\shortstack{\includegraphics[width=0.15\linewidth]{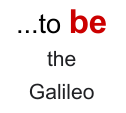}}&
\shortstack{\includegraphics[width=0.15\linewidth]{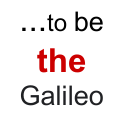}}&
\shortstack{\includegraphics[width=0.15\linewidth]{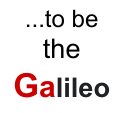}}\\[1pt]
\end{tabular}
}
    \caption{Comparison between different styles in the same input audio.}
    \label{fig:fixedAudio}
\end{figure}

Additionally, in Figure~\ref{fig:fixedAudio}, we also show the comparison between different styles when they are encoded in one input audio to generate talking heads. Note that, in this case, different input images are used to verify the synthesis effectiveness of our method. Although different styles are encoded into different images to generate different talking heads, the animation is realistic and the performance of lip-synchronization is well-reserved. More illustrative results can be found in our demonstration video.

\subsection{Quantitative Evaluation}

\subsubsection{Evaluation Metric}

\textbf{Metrics to evaluate 2D talking head results.} We use six different metrics to evaluate how good and natural the animation of the generated talking head is. They are: Cumulative Probability Blur Detection (CPBD)~\cite{vougioukas2019realistic}, Landmark Distance (D-L)~\cite{zhou2020makelttalkMIT}, Landmarks Distance around the Mouth (LMD), Landmark Velocity difference (D-V)~\cite{zhou2020makelttalkMIT}, Difference in the open mouth area (D-A)~\cite{zhou2020makelttalkMIT}.

\textbf{Metric to evaluate style transfer.} The aforementioned metrics such as D-L, LMD, D-V, and D-A require the ground truth and the predicted sample to be synchronized with each other to measure the accuracy of the generated talking head. They also consider all frames to be equally important and calculate the accuracy by averaging all frames. Therefore, these metrics do not take into account the style information, which has more temporal dynamics and special facial expressions during a short duration.
To evaluate style transfer results efficiently, we introduce three new following metrics. 

\textit{Style-Aware Landmarks Distance} (SLD): To evaluate the style information encoded in a generated talking head, we design a metric called Style-Aware Landmarks Distance (SLD). This metric calculates the accuracy of mouth, eyes, head pose shapes between a chunked window of style reference and a chunked window of corresponding talking head animation. Lower is better. 
Let's assumed that a style reference video with $N_s$ frames is split into multiple temporal periods of $\rm F$ frames (window size), i.e., style reference windows $\rm W_s = \left(w_s^{(0:F)}, w_s^{(v:F+v)}, w_s^{(2v:F+2v)},\cdots, w_s^{(\kappa v:F +\kappa v)}\right)$, with $\rm w_s^{(i:F+i)}$ being the frames from $\rm i$\textit{-th} to $\rm (F+i)$\textit{-th}
of the reference video, $\rm v$ is the stride, and $\rm \kappa = \lfloor(N_s - F) / v\rfloor$.
Similar to the reference video, we chunk the generated animation video
into smaller chunked windows $\rm W_a = \Bigl(w_a^{(0:F)}, w_a^{(v:F+v)}, w_a^{(2v:F+2v)}, \cdots , \allowbreak w_a^{(\kappa v:F+ \kappa v)}\Bigr)$. The SLD is then calculated with the core is the D-L metric as:
\begin{equation}
    \label{eq:SLD}
    \rm{SLD} = \frac{1}{\rm \vert W_s\vert}\sum_{\rm w_s \in \rm W_s} \left(\underset{\rm w_a \in \rm W_a}{\rm{min}} \left(\rm{D\rm{-}L}\left(\rm w_s, \rm w_a\right)\right)\right) 
\end{equation} 
where $\rm{D\rm{-}L}$ is the Landmark Distance metric~\cite{zhou2020makelttalkMIT}.

Similarly, we calculate the 
\textit{Style-Aware Landmarks Velocity Difference} (SLV) and \textit{Style-Aware Mouth Area Difference} (SMD) as follow:

\begin{equation}
    \label{eq:SLV}
    \rm{SLV} = \frac{1}{\rm \vert W_s\vert}\sum_{\rm w_s \in \rm W_s} \left(\underset{\rm w_a \in \rm W_a}{\rm{min}} \rm{D\rm{-}V}\left(\rm w_s, \rm w_a\right)\right)
\end{equation} 
where ${\rm D\rm{-}V}$ is the Landmark Velocity difference metric~\cite{zhou2020makelttalkMIT}.

\begin{equation}
    \label{eq:SMD}
    \rm{SMD} = \frac{1}{\rm \vert W_s\vert}\sum_{\rm w_s \in \rm W_s} \left(\underset{\rm w_a \in \rm W_a}{\rm{min}} \left(\rm LMD\left(\rm w_s, \rm w_a\right)\right)\right)
\end{equation} 
where ${\rm LMD}$ is Landmarks Distance around the Mouth~\cite{chung2016lip}.

To robustly compare our results, we construct a grid of window size $\rm {F}=\{1,2,...,100\}$ and stride $\rm{v}=\{1,2,...,20\}$ and then compute the above metrics, i.e., SLD, SLV, and SMD, for each element on the grid. The final value is then calculated as the average over the grid of all computed metric values corresponding to each window size and stride setting.
In all metrics for evaluating style transfer, the function $\rm{min}(\cdot)$ is used to search for the best matched local window, i.e., a temporal period of frames considered to contain the best-matched style information. 
 In this way, our style metrics can take into account the temporal, and then support validating the style information encoded in it. Note that, we assume one video would have only one style when applying our proposed metrics.

\subsubsection{Dataset}  Since our method focus on learning different character styles in different circumstances, we evaluate and benchmark our results in the RAVDESS dataset~\cite{livingstone2018ryersonRAVDESS}. The RAVDESS is a validated multimodal database of emotional speech and song, which is suitable and challenging to validate our method and different baselines. Note that, we only use this dataset for benchmarking  to avoid training bias.

\subsubsection{2D Talking Head Generation Results}

\begin{table}[!ht]
\centering
\caption{Result of different 2D talking head generation methods.}

\setlength{\tabcolsep}{0.3 em} 
{\renewcommand{\arraystretch}{1.5}
\begin{tabular}{c|c|c|c|c|c}
\hline
\multirow{2}{*}{\textbf{Methods}} & \multicolumn{5}{c}{\textbf{Metrics}}  \\ \cline{2-6} 
                                   & \multicolumn{1}{c|}{\textbf{CPBD$\uparrow$}} & \multicolumn{1}{c|}{\textbf{LMD$\downarrow$}} & \textbf{D-L$\downarrow$} & \textbf{D-V$\downarrow$} & \textbf{D-A$\downarrow$}
                        \\ \hline
Ground Truth    	&0.28	&0.00	&0.00 \% &0.00 \% &0.00 \%                 \\ \hline
MIT~\cite{zhou2020makelttalkMIT}  &0.18	&2.28	&2.78\%	&0.88\%	 &14.52\%                \\
PCT~\cite{zhou2021pose}   	&0.09	&3.22	&3.27\%	&0.86\%	&36.84\%                   \\
LSP~\cite{lu2021liveLSP}   	&0.20	&3.29	&5.43\%	&0.85\%	&30.65\%               \\
AD-NERF~\cite{guo2021adneft} 	&0.21	&2.43 &2.67\% &0.85\% &13.34\%	                  \\ \hline \hline
Ours     &\textbf{0.26}	&\textbf{1.83}  &\textbf{2.65\%}	&\textbf{0.83\%}	&\textbf{10.53\%}               \\\hline
\end{tabular}
}
\vspace{2ex}

\vspace{-2ex}
\label{tab:baseline}
\end{table}

Table~\ref{tab:baseline} shows the 2D talking head result comparison between our method and recent baselines, including~\cite{zhou2020makelttalkMIT, zhou2021pose, lu2021liveLSP, guo2021adneft}. From Table~\ref{tab:baseline}, we can see that our method outperforms recent state-of-the-art approaches by a large margin. In particular, our method achieves the highest accuracy in CPBD, LMD, D-L, D-V, and D-A metrics. These results show that our method successfully renders the 2D talking head and increases the quality of the rendered results. Overall, our method can increase the sharpness of the head (identified by CPBD) metric, while generating natural facial motion (identified by LMD, D-L, D-V, and D-A metric). 

\subsubsection{Style Comparison}

Table~\ref{tab:style_quantitative} shows the comparison between our method and four baselines~\cite{zhou2020makelttalkMIT, zhou2021pose, lu2021liveLSP, guo2021adneft} in terms of style transfer. Three designed metrics (SLD, SLV, and SMD) are used for evaluation and benchmarking.
The results show that our method outperforms others by a large margin in all three  metrics, which suggests that our method effectively captures style information from the style reference and successfully transfer it to the target image. 

\begin{table}[!ht]
\centering
\caption{Result comparison in terms of style transfer between different 2D talking head generation methods.}

\setlength{\tabcolsep}{1.0 em} 
{\renewcommand{\arraystretch}{1.5}
\begin{tabular}{c|c|c|c}
\hline
\multirow{2}{*}{\textbf{Methods}} & \multicolumn{3}{c}{\textbf{Metrics}} \\ \cline{2-4} 
 & \textbf{SLD$\downarrow$} & \textbf{SLV$\downarrow$} & \textbf{SMD$\downarrow$} \\ \hline
MIT~\cite{zhou2020makelttalkMIT} &3.00  &0.94  &5.03   \\ 
PCT~\cite{zhou2021pose} &3.58  &0.93  &7.28  \\ 
LSP~\cite{lu2021liveLSP} &5.40  &0.91  &6.89  \\ 
AD-NEFT~\cite{guo2021adneft} &4.69 &0.92  &5.48  \\ \hline \hline
Ours &\textbf{2.84}  &\textbf{0.89}  &\textbf{4.26} \\ \hline
\end{tabular}
}
\vspace{2ex}

\vspace{-2ex}
\label{tab:style_quantitative}
\end{table}

\begin{figure}[t]
   \centering
\resizebox{\linewidth}{!}{
\setlength{\tabcolsep}{2pt}
\begin{tabular}{ccccc}

\shortstack{\rotatebox[origin=l]{90}{\hspace{0.35cm}
}}&
\shortstack{\includegraphics[width=0.33\linewidth]{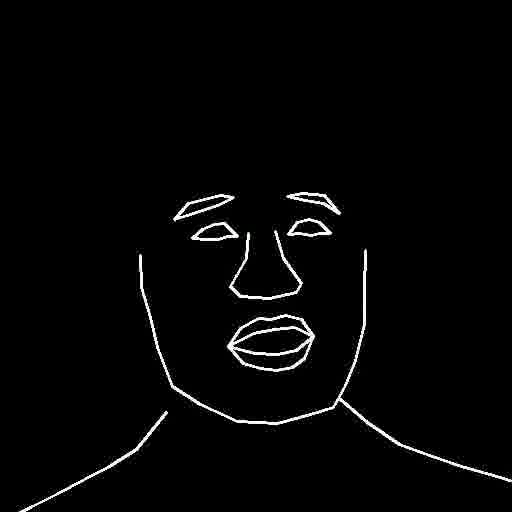}}&
\shortstack{\includegraphics[width=0.33\linewidth]{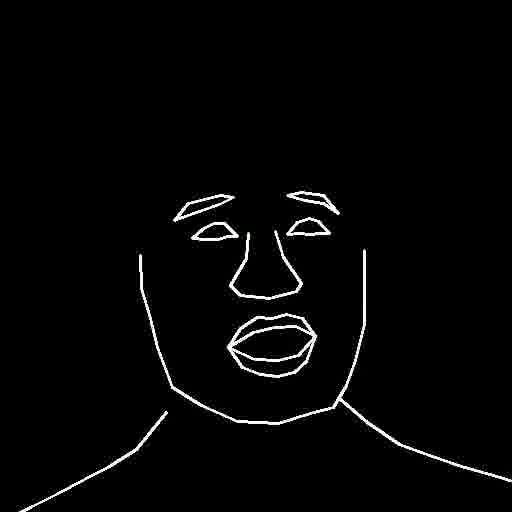}}&
\shortstack{\includegraphics[width=0.33\linewidth]{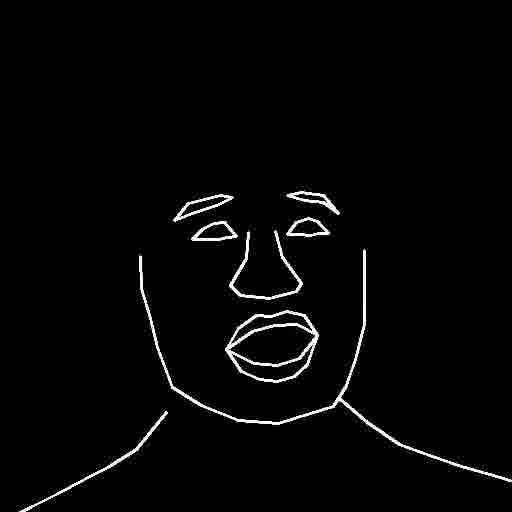}}&
\shortstack{\includegraphics[width=0.33\linewidth]{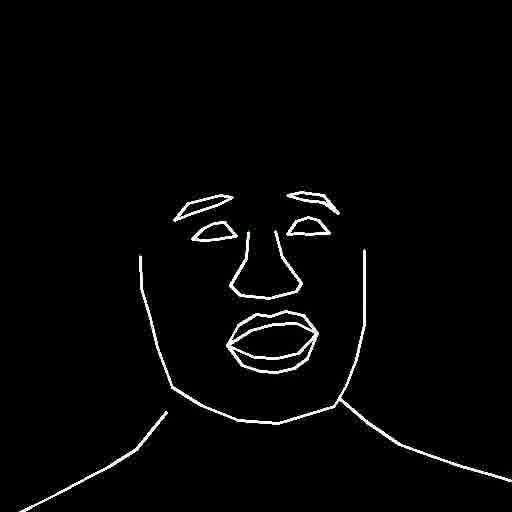}}\\[1pt]
\rotatebox[origin=l]{90}{\hspace{0.4cm} 
}&
\shortstack{\includegraphics[width=0.33\linewidth]{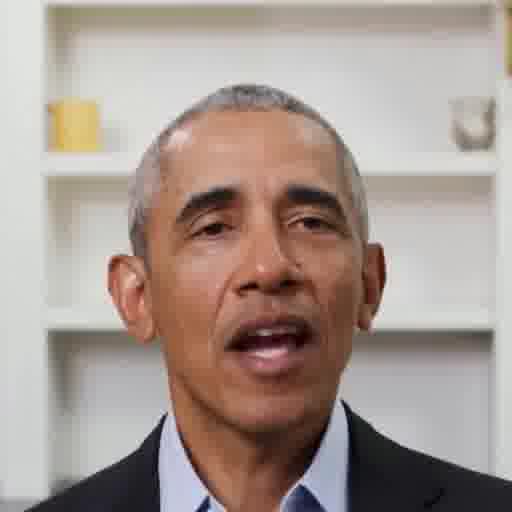}}&
\shortstack{\includegraphics[width=0.33\linewidth]{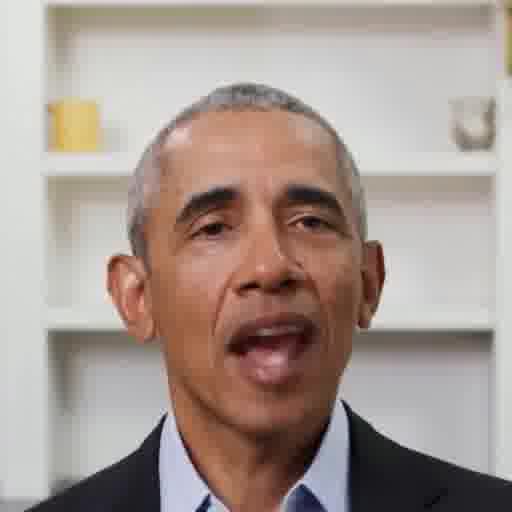}}&
\shortstack{\includegraphics[width=0.33\linewidth]{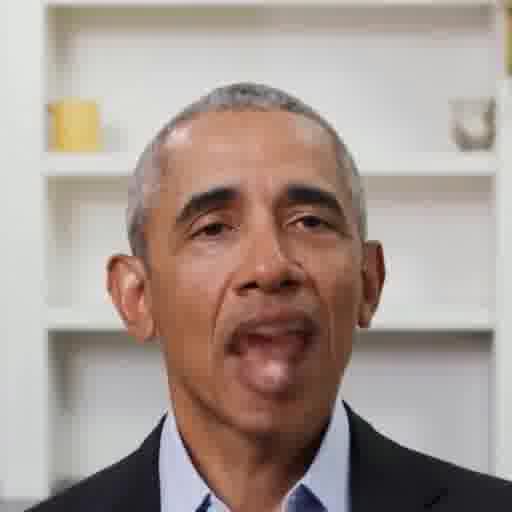}}&
\shortstack{\includegraphics[width=0.33\linewidth]{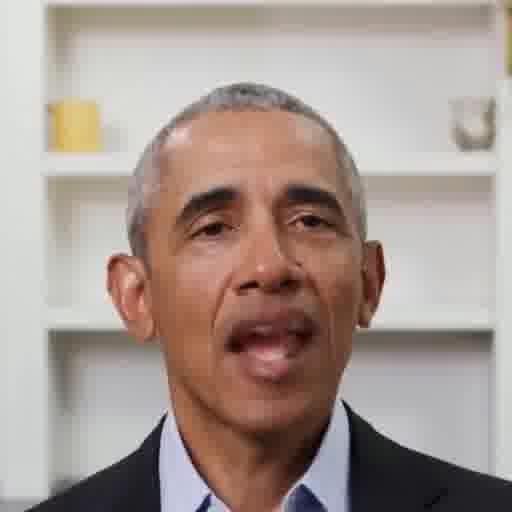}}\\[1pt]
\rotatebox[origin=l]{90}{\hspace{0.55cm} 
}&
\shortstack{\includegraphics[width=0.33\linewidth]{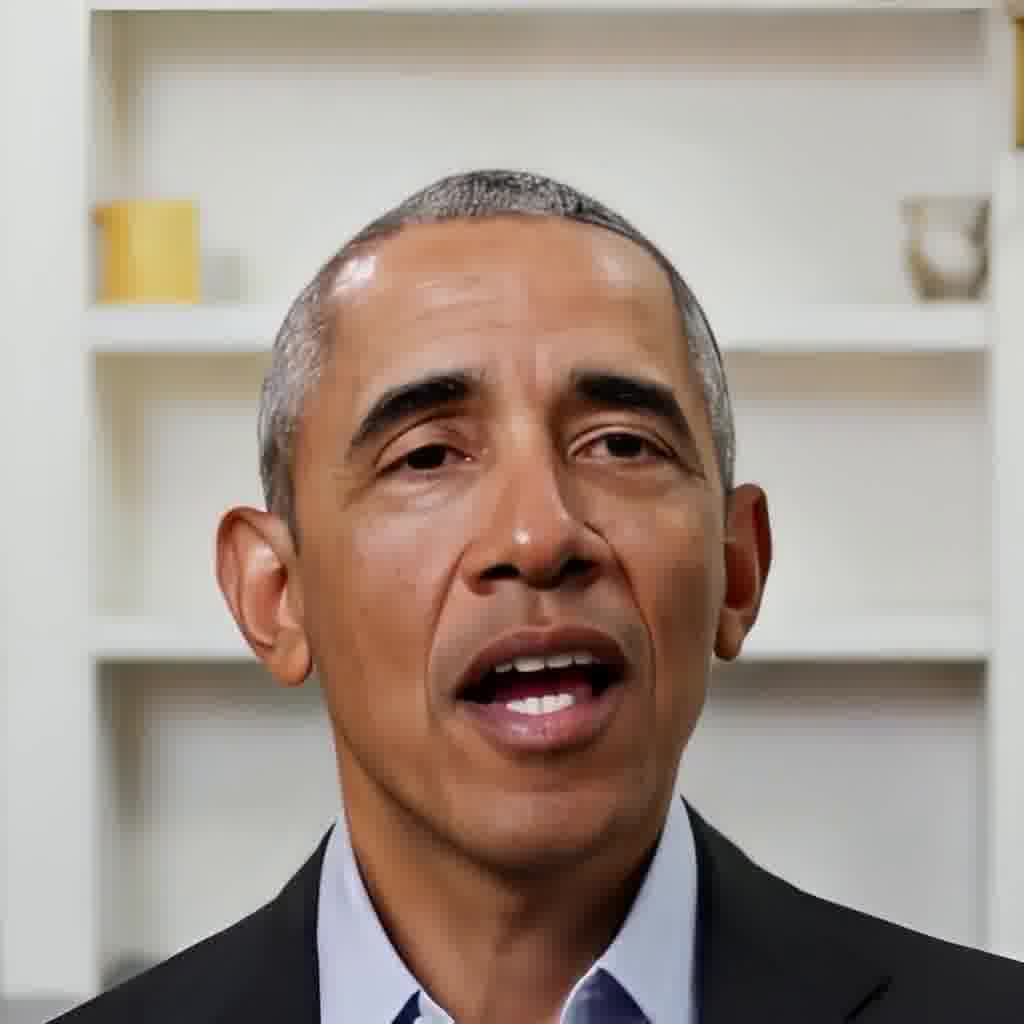}}&
\shortstack{\includegraphics[width=0.33\linewidth]{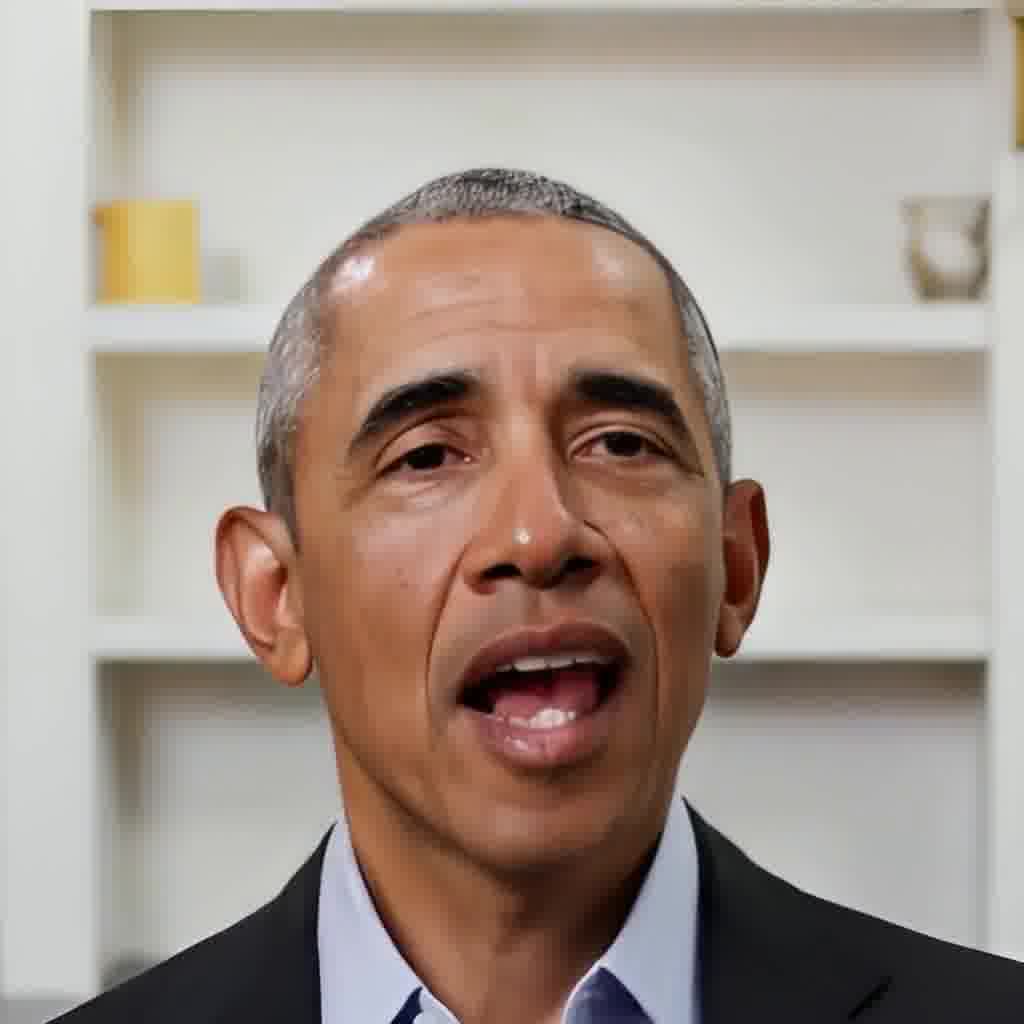}}&
\shortstack{\includegraphics[width=0.33\linewidth]{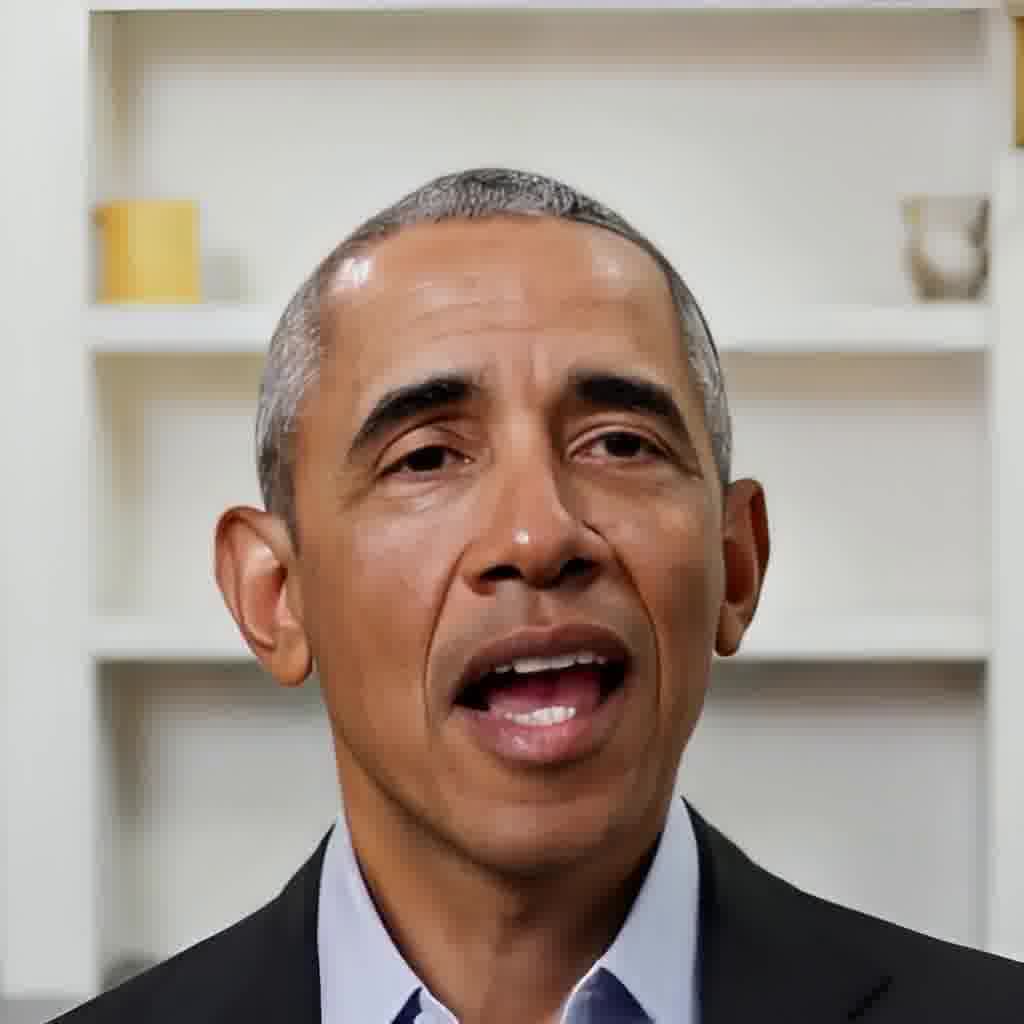}}&
\shortstack{\includegraphics[width=0.33\linewidth]{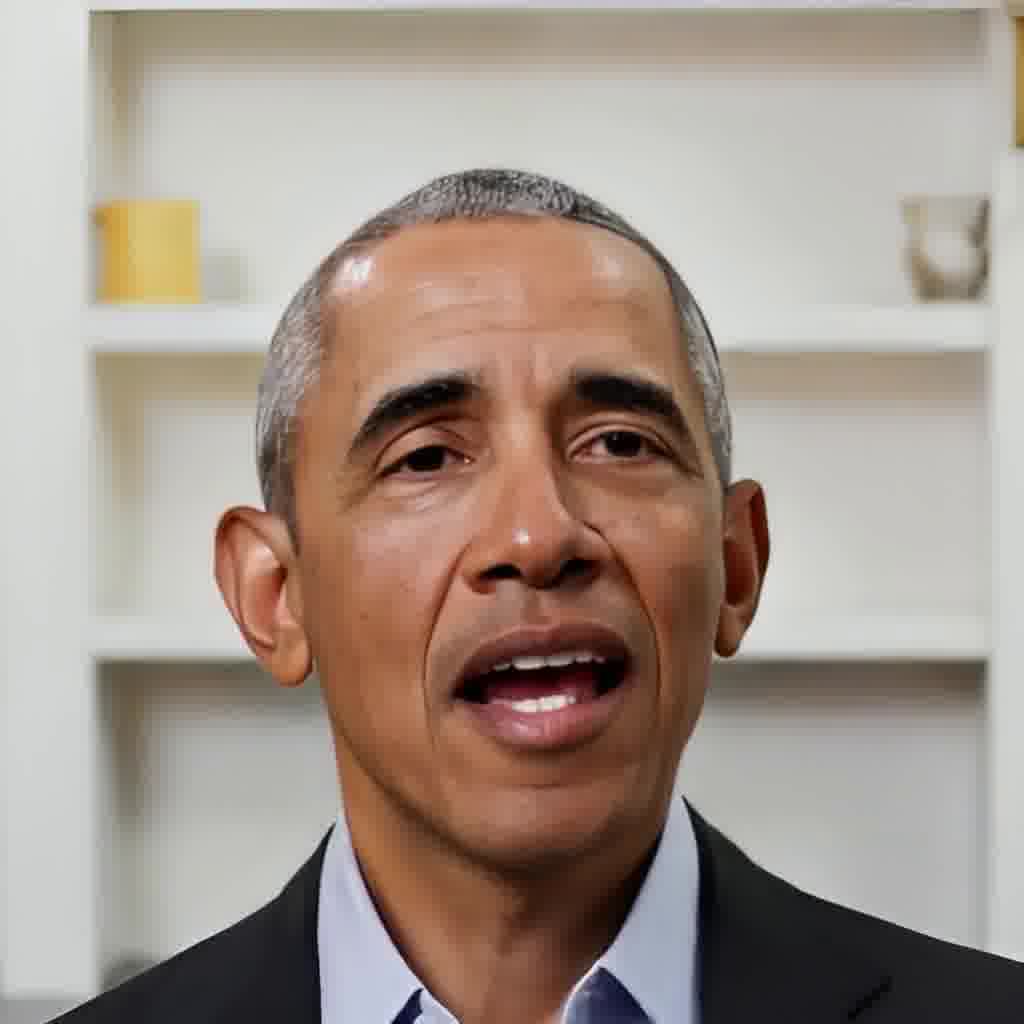}}\\[1pt]
\end{tabular}
}
    \caption{Effectiveness of ISP and Style-Aware loss. First row: The facial map; Second row: the 2D talking head without ISP and Style-Aware loss; Third row: the 2D talking head with ISP and Style-Aware loss.}
    \label{fig:AblStyle}
\end{figure}

\subsection{Intermediate Style Pattern Analysis}
In Figure~\ref{fig:AblStyle}, we investigate the effectiveness of our Style-Aware Generator when using the ISP. This figure illustrates how the ISP and the Style-Aware loss (Equation~\ref{eq_style_aware_loss}) provide meaningful information to generate better photo-realistic results. Overall, we observe that the rendered frames using the ISP have more realistic and detailed faces, especially around the mouth and the eye of the character.

Specifically, in some exceptional circumstances (e.g., in the \texttt{opera} singing style, the mouth of the character is widely opened), the face motions contain the unique shape of landmarks which is rare or not well captured in the training data (see the first row of Figure~\ref{fig:AblStyle}). As a consequence, the Style-Aware Generator, which is responsible for rendering 2D motions, cannot handle this problem. Hence, the rendered faces are not realistic and have blurry and ghosting effects (i.e., the second row of Figure~\ref{fig:AblStyle}). To address this problem, the ISP and Style-Aware loss can be utilized to mitigate these effects. The ISP can give useful information about the appearance of the stylized motion and help the image generator focus on critical visual cues. As the result, our designed Style-Aware Generator produces better photo-realistic results (see the third row of Figure~\ref{fig:AblStyle}).

\begin{figure}[!t]
    \centering
    \includegraphics[width=0.48\textwidth, keepaspectratio=true]{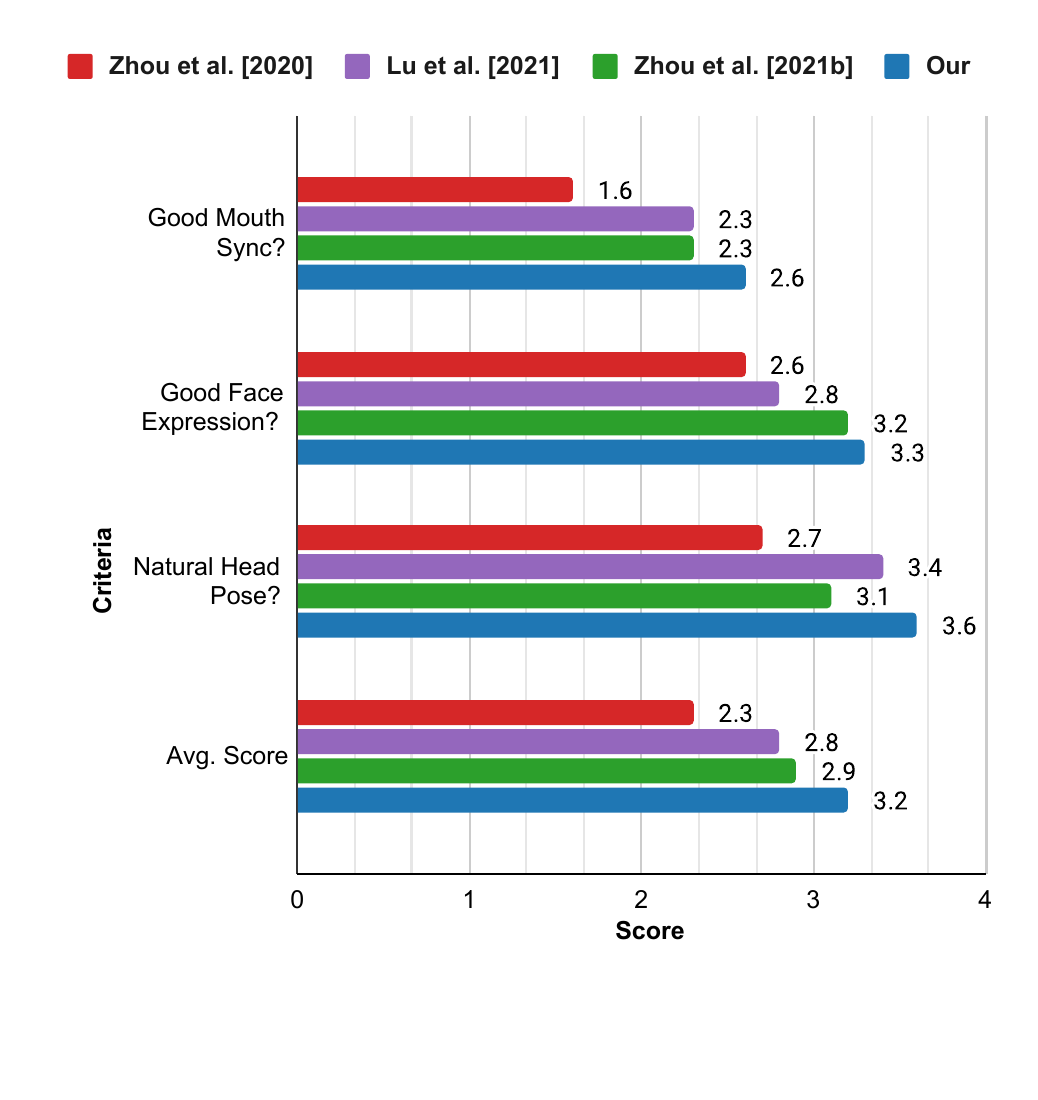}
    \caption{User study results of three criteria forming natural talking heads.}
    \label{fig:UserStudy1}
\end{figure}

\subsection{User Study}
\label{subsec:UserStudy}
We further conduct user studies to verify our style-aware talking head generation method. We set up three user studies and recruit 56 people with different backgrounds for our experiment. In the first and second studies, we compare the naturality of talking heads and how well the styles are transferred between our work and recent work PCT~\cite{zhou2021pose}, MIT~\cite{zhou2020makelttalkMIT}, LSP~\cite{lu2021liveLSP}. In the third study, we verify the robustness of our proposal in transferring personalized style. Note that, to achieve fair judgment, the users only see the output images/videos, and not the name of any methods in all studies.

\subsubsection{Natural Talking Head Animation Study}
Throughout this study, our app will play one video at a time in a randomized order, and each participant will be asked to rate the video based on three statements: \textit{(i)} Is the mouth of the talking person synchronized with the corresponding audio? \textit{(ii)} Is the expression of the face appropriate for the audio? And \textit{(iii)} Is the head motion natural?. We set a score band between 1 to 4 (4-yes, 3-yes but some parts of the video are not good enough, 2-no but some parts of the video are pleased, 1-no). We make 30 videos for each method. 
All of the mentioned videos have inputs unseen in the training set. Figure~\ref{fig:UserStudy1}  demonstrates the average scores of different methods on three  questions. 

It can be seen that our method achieves the best results across all three questions. Our approach received the highest score of $2.6$ for the first question, indicating that our findings have the best mouth synchronization outcomes,  especially with challenging cases like \texttt{rap} or \texttt{opera} style. We believe that the Style-Aware Generator is successfully trained to generalize and capture unique mouth shapes in challenging cases. However, based on the user feedback, we note that there is room for further improvement. In the meanwhile, other questions show that our method captures the facial expression more effectively and provides more realistic head motions than other recent methods.

\begin{figure}[!t]
    \centering
    \includegraphics[width=0.48\textwidth, keepaspectratio=true]{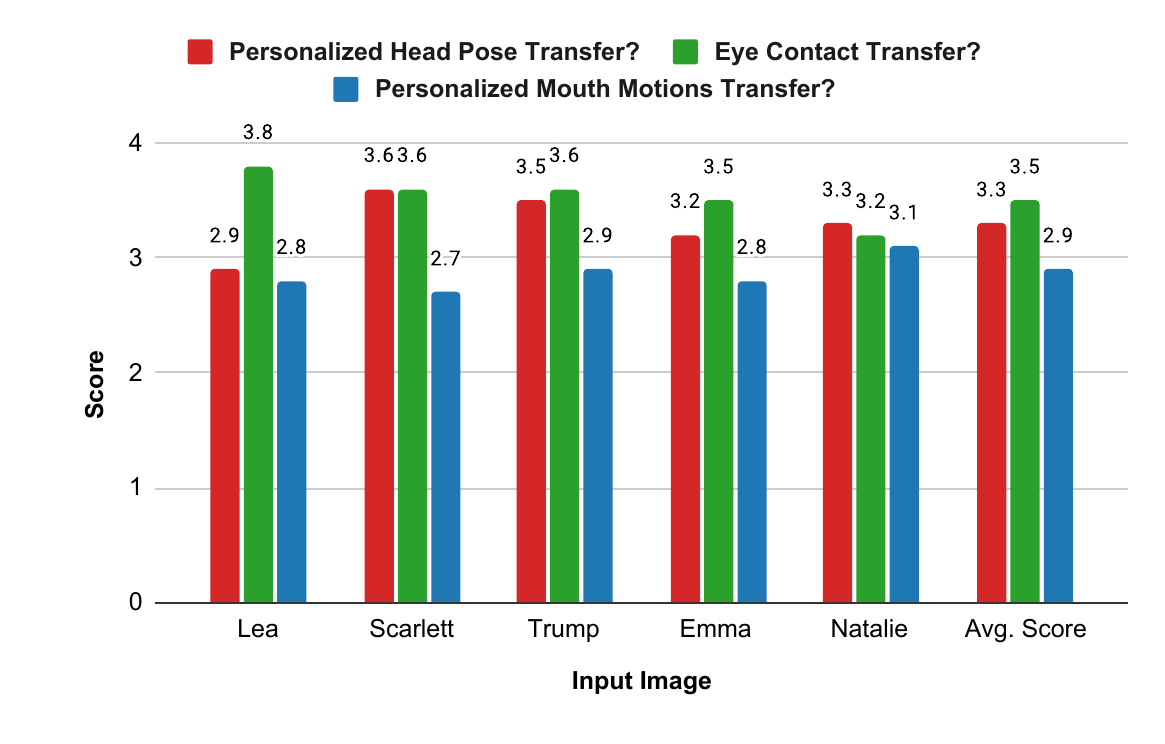}
    \caption{User study results which verify the effectiveness of the style transfer process when different styles are transferred on fixed input images to generate talking heads.}
    \label{fig:UserStudy3}
\end{figure}

\subsubsection{Style Transfer: Robustness Study} This study analyses different characteristics of the style transfer results in our method. In this study, our app shows an input image of a character, a style reference video, an audio, and the corresponding generated talking head video. 
To be more specific, we use a single image from each of the five persons in this study, which are Lea, Scarlett, Trump, Emma, and Natalie. We also collected nine talking/singing clips on the internet to use as the style reference videos. For each style reference, we first apply our transferring procedure and leverage the learned model to generate 10 style transferred talking head animations corresponding to 10 audio sequences. 
Each participant is asked if the style reference video and the style transferred 2D talking head animations represent a similar talking style.
There are three questions to verify whether the style is successfully transferred: \textit{(i)} How successful is the style of head pose transferred?, \textit{(ii)} How successful is the style of eye contact transferred?, and \textit{(iii)} How successful is the style of mouth motions transferred?
The band score is similar to the previous user study.

Figure~\ref{fig:UserStudy3} demonstrates the scores of our methods on five fixed images that have talking heads generated from different style references. Although our method achieves reasonable scores in head pose transfer and eye contact criteria (3.3 and 3.5 on average, respectively), the score on the mouth motion transfer criteria is lower (2.9 in average). 
This result confirms that, although our method can capture the unique shape of mouth landmarks in style references and transfer it into input images, this task is not trivial and needs more improvement in the future.

\section{Conclusion}
\label{Sec:Conclusion}

We have proposed a deep learning framework that creates 2D talking heads from the input audio. Besides an audio stream and an image, our framework utilizes a set of reference frames to learn the character style. Our proposed method can successfully extract and transfer the reference style to the output 2D talking head animation. In practice, our method successfully creates photo-realistic and high fidelity animations. Furthermore, apart from the normal talking style, our method can also work well with challenging singing styles such as \texttt{ballad}, \texttt{opera}, and \texttt{rap}, which requires an adaptable translation module to generate detailed and accurate animations. The intensive experiments show that our talking head results outperform other recent state-of-the-art approaches qualitatively and quantitatively. Our framework can be used in different applications such as dubbing, video conferencing, and virtual avatar assistant.

\ifCLASSOPTIONcaptionsoff
  \newpage
\fi

\bibliographystyle{IEEEtran}
\bibliography{IEEEabrv}

\end{document}